\documentclass[journal,11pt, onecolumn]{IEEEtran}

\date{} 

\let\labelindent\relax
\usepackage{enumitem}

\usepackage{url}

\usepackage{hyperref} 

\usepackage{blindtext}
\usepackage[dvipsnames]{xcolor}
\usepackage{subfig}
\usepackage{amsmath}
\usepackage{mathrsfs}
\usepackage{caption}
\usepackage{amssymb}
\usepackage{mathabx}
\usepackage[ruled,linesnumbered]{algorithm2e}
\DeclareMathAlphabet\mathbfcal{OMS}{cmsy}{b}{n}
\usepackage{multirow}
\usepackage{hyperref}
\usepackage{tablefootnote}
\usepackage{graphicx}

\newsavebox{\measurebox}

\usepackage{tikz}
\usetikzlibrary{mindmap,trees,shadows,backgrounds}
\tikzstyle{every annotation} = [fill=blue!20]

\usepackage[noadjust]{cite}

\ifCLASSINFOpdf
\else
\fi
\begin{document}

\onecolumn 

\begin{description}[labelindent=0cm,leftmargin=4cm,style=multiline]

\item[\textbf{Citation}]{G. AlRegib, M. Deriche, Z. Long, H. Di, Z. Wang, Y. Alaudah, M. Shafiq, and M. Alfarraj, “Subsurface structure analysis using computational interpretation and learning: A visual signal processing perspective,” IEEE Signal Processing Magazine, vol. 35, no. 2, pp. 82-98, Mar. 2018.}
\\
\item[\textbf{DOI}]{\url{https://doi.org/10.1109/MSP.2017.2785979}}
\\
\item[\textbf{Review}]{Date of publication: 12 March 2018}
\\
\item[\textbf{Data and Codes}]{\url{https://ghassanalregib.com/}}
\\
\item[\textbf{Bib}] {@ARTICLE\{8312469, \\
author=\{G. AlRegib and M. Deriche and Z. Long and H. Di and Z. Wang and Y. Alaudah and M. A. Shafiq and M. Alfarraj\}, \\
journal=\{IEEE Signal Processing Magazine\}, \\
title=\{Subsurface Structure Analysis Using Computational Interpretation and Learning: A Visual Signal Processing Perspective\}, \\
year=\{2018\}, \\
volume=\{35\}, \\
number=\{2\}, \\
pages=\{82-98\},\\ 
doi=\{10.1109/MSP.2017.2785979\}, \\
ISSN=\{1053-5888\}, \\
month=\{March\}\}
} 
\\

\item[\textbf{Copyright}]{\textcopyright 2018 IEEE. Personal use of this material is permitted. Permission from IEEE must be obtained for all other uses, in any current or future media, including reprinting/republishing this material for advertising or promotional purposes, creating new collective works, for resale or redistribution to servers or lists, or reuse of any copyrighted component of this work in other works.}
\\
\item[\textbf{Contact}]{\href{mailto:zhiling.long@gatech.edu}{zhiling.long@gatech.edu}  OR \href{mailto:alregib@gatech.edu}{alregib@gatech.edu}\\ \url{https://ghassanalregib.com/} \\ }
\end{description}

\thispagestyle{empty}
\newpage
\clearpage
\setcounter{page}{1}

%
\title{Subsurface Structure Analysis using Computational  Interpretation and Learning: A Visual Signal Processing Perspective}
%
%
%
\author{Ghassan AlRegib,~\IEEEmembership{Senior Member,~IEEE,}
Mohamed Deriche,~\IEEEmembership{Senior Member,~IEEE,}
Zhiling Long,~\IEEEmembership{Member,~IEEE,}
Haibin Di,~\IEEEmembership{Member,~IEEE,}
Zhen Wang,~\IEEEmembership{Student Member,~IEEE,}
Yazeed Alaudah,~\IEEEmembership{Student Member,~IEEE,}
Muhammad A. Shafiq,~\IEEEmembership{Student Member,~IEEE,} and
Motaz Alfarraj,~\IEEEmembership{Student Member,~IEEE}

\thanks{G. AlRegib is with the School of Electrical and Computer Engineering, Georgia Institute of Technology, Atlanta,
GA 30332-0250, USA e-mail: alregib@gatech.edu}
\thanks{M. Deriche is with the Electrical Engineering Department, King Fahd University of Petroleum \& Minerals, Saudi Arabia}
\thanks{Z. Long, H. Di, Z. Wang, Y. Alaudah, M. Shafiq, and M. Alfarraj are with the School
of Electrical and Computer Engineering, Georgia Institute of Technology, Atlanta, GA 30332-0250, USA}
}

%
%

\markboth{IEEE Signal Processing Magazine}%
{AlRegib \MakeLowercase{\textit{et al.}}: Subsurface Structure Analysis using Computational  Interpretation and Learning}
%



\maketitle

\begin{abstract}

Understanding Earth's subsurface structures has been and continues to be an essential component of various applications such as environmental monitoring, carbon sequestration, and oil and gas exploration. By viewing the seismic volumes that are generated through the processing of recorded seismic traces, researchers were able to learn from applying advanced image processing and computer vision algorithms to effectively analyze and understand Earth's subsurface structures. In this paper, first, we summarize the recent advances in this direction that relied heavily on the fields of image processing and computer vision. Second, we discuss the challenges in seismic interpretation and provide insights and some directions to address such challenges using emerging machine learning algorithms. 

\end{abstract}


%
\IEEEpeerreviewmaketitle

\section{Introduction}

Seismic interpretation is a critical process in subsurface exploration. The process aims at identifying structures or environments of significant importance in diverse applications. For example, for oil and gas exploration, a successful interpretation can help identify structures (such as faults, salt domes, and horizons, etc.) that are indicators of potential locations of reservoirs. Reliable delineation of natural faults and fractures will help predict crucial geologic deformation events and monitor potential earthquakes. In environmental engineering, capture and storage of carbon dioxide is substantially dependent on accurate subsurface interpretation to find a good geologic trap with minimum leakage probability. When subsurface structures are the primary interest in seismic interpretation, it is often called \textit{structural interpretation}. However, when environments such as patterns of deposition are of concern, it is then called \textit{stratigraphic interpretation}. In this paper, we focus on structural interpretation.

Typically, interpreters first assume a geological model based on the analysis of various attributes of the data, in addition to the geological history of the region. They then manually segment the data volume into sub-volumes based on the dominant structures contained within each sub-volume. Thereafter, interpreters perform interpretation to delineate various subsurface structures of interest. Based on the results, the geological model is modified, and the process is repeated until the interpreters converge on a geologically plausible and reasonably accurate model. Such process is increasingly becoming more time consuming and labor intensive with the explosive growth in data. Seismic surveys have grown over the years in both complexity and data sizes. Data sizes from such surveys have increased from hundreds of megabytes of the first 3D seismic survey collected in 1970's to thousands of gigabytes or even terabytes nowadays. To tackle this challenging situation, automated or semi-automated interpretation through computational algorithms have been investigated.

For years, image  processing  theories  and  algorithms  have  been employed  in  seismic  interpretation  and made essential contributions to the field. Given the richness and fast progress in image processing as well as related areas of computer vision and machine learning, there remains a significant room to explore their impact on achieving a higher level of automation in seismic interpretation. Equally important, we believe that seismic interpretation is a challenging problem that has and continues to impact the advancement of image processing and related fields.

\begin{figure}[h]
\begin{center}
\includegraphics[width= 6 in]{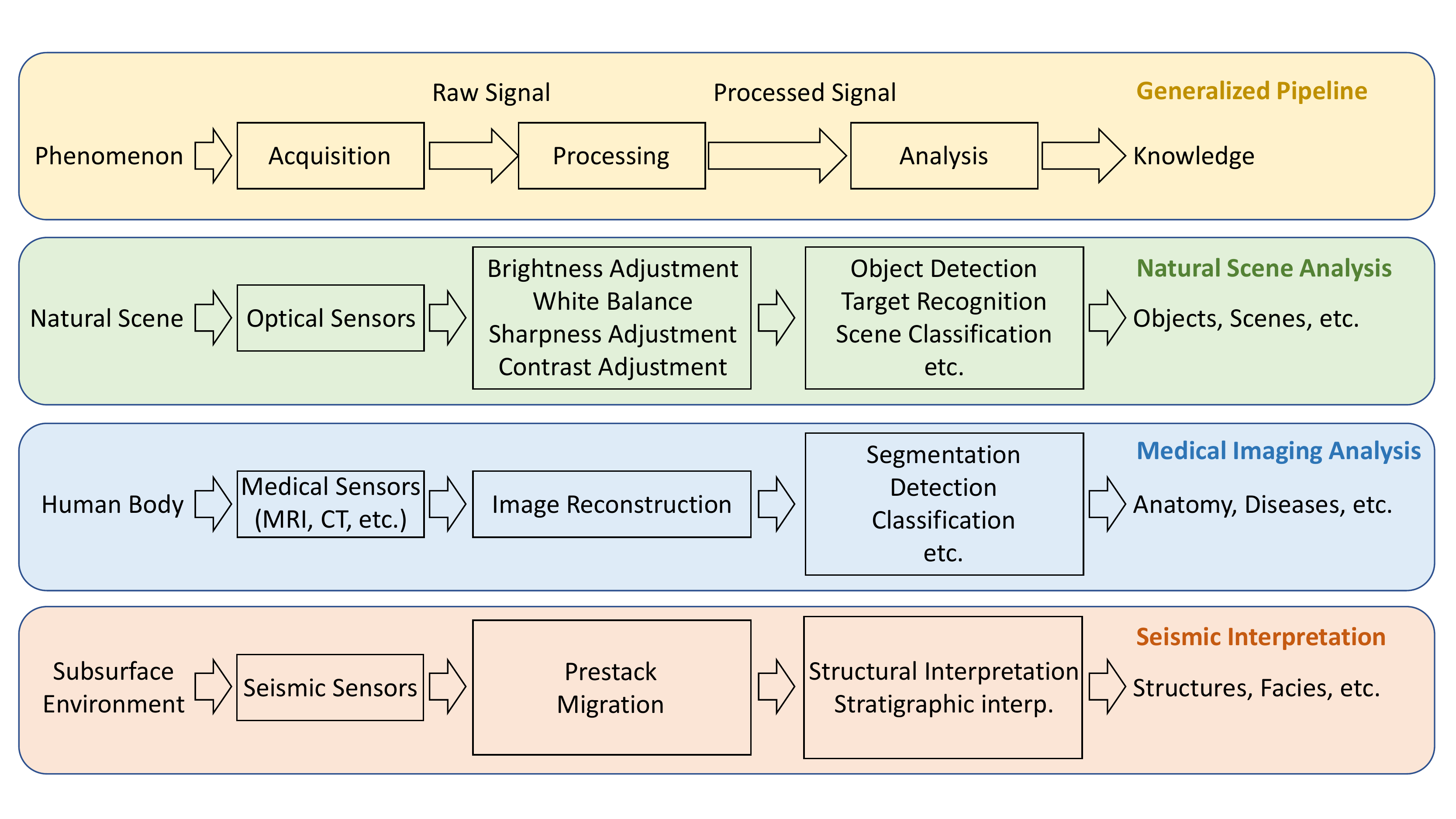}
\end{center}
\caption
{Analogy between seismic interpretation, natural scene analysis, and medical imaging analysis.}
\label{fig:analogy}
\end{figure}

Although a deep understanding of, and training in geophysics have been considered prerequisites for developing effective computational interpretation algorithms, the recent progress in machine learning has shined a new light towards the roles that this field can play in domain-specific problems. We believe that, in essence, seismic interpretation fits very well into a generalized pipeline for knowledge discovery through imaging applications, in a similar way to natural scene analysis and medical imaging analysis do. This is illustrated in Fig.~\ref{fig:analogy}, where all these processes share a similar pipeline that consists of modules for acquisition, processing, and analysis. For the seismic case, the acquisition is fulfilled by sensing devices such as geophones, for land surveys, which collect seismic waves reflected by subsurface structures, caused by a controlled seismic source of energy (e.g., a seismic vibrator). The acquired signals then go through advanced signal- and wave-equation-based processing (normally prestack migration) to yield data for the stage of analysis, where seismic interpretation takes place. Such a framework aims at transforming a certain physical phenomenon into signals that can be analyzed to understand that phenomenon. The differences lie in the methodologies within each module and the underlying designs that accommodate the physical principles behind the phenomena being observed, which impose certain constraints on the properties of the signals. 

The focus of this article is to present the seismic interpretation in the framework of the generalized pipeline for imaging data analysis where both human visual system-based models and learning-based models have the potential to assist in discovering underlying structures that may not have been discovered otherwise. We discuss subsurface event detection and tracking as well as labeling and classification. We also discuss recent trends that aim at training machine learning models to understand and reveal various subsurface structures. More specifically, we frame the problem of delineation as a tracking problem, using video coding as an example (as presented in Sec. III-C). Moreover, the seismic interpretation pipeline can be re-designed as a labeling problem, modeling after the natural scene labeling problem (see Sec. IV). Furthermore, incorporating an understanding of the human visual system (HVS) can facilitate automating the process, mimicking the process of a human interpreter perceiving and analyzing a large volume of a post-stack seismic dataset. An example of this is to design interpretation algorithms based on attention models as in the HVS (see Sec. V).

We hope that this article will make the challenges found in seismic interpretation more accessible to the signal processing community. The start-up tools and datasets provided in the article will enable the interested individuals to jump start in this domain, especially if they do not have a background in geophysics. Furthermore, analyzing the seismic volume from the HVS point of view can unveil excellent opportunities that may not exist otherwise. Finally, solving challenges that are rooted in the unique characteristics of seismic data will help advance the corresponding signal processing and machine learning theories and algorithms as well.

\section{Subsurface Structures and Datasets}

Before proceeding to the review of common interpretation tasks, in this section, we provide a brief introduction to several subsurface structures of interest, followed by a list of datasets that are commonly used for developing and testing the algorithms. Throughout this paper, we will use oil and gas exploration as the example to illustrate subsurface understanding through seismic interpretation for a number of reasons. Compared to other applications, data is more abundantly available with oil and gas exploration. Also, there is a rich literature on this subject with various contributions that can provide a benchmark. Furthermore, the depth of the imaged subsurface is deeper in the application of consideration, and that adds to the complexity of the structure to be interpreted. Therefore, the structures and datasets introduced here are typically related to hydrocarbon reservoir identification and characterization, although some of them can also be used for other applications such as earthquake monitoring and environmental engineering, which is a common practice in the community. Nevertheless, all reviewed and proposed algorithms are applicable in all applications that image the subsurface to locate certain structures. 

\subsection{Common Subsurface Structures}
\begin{figure}[ht!]
	\centering
	\includegraphics[width = 4in]{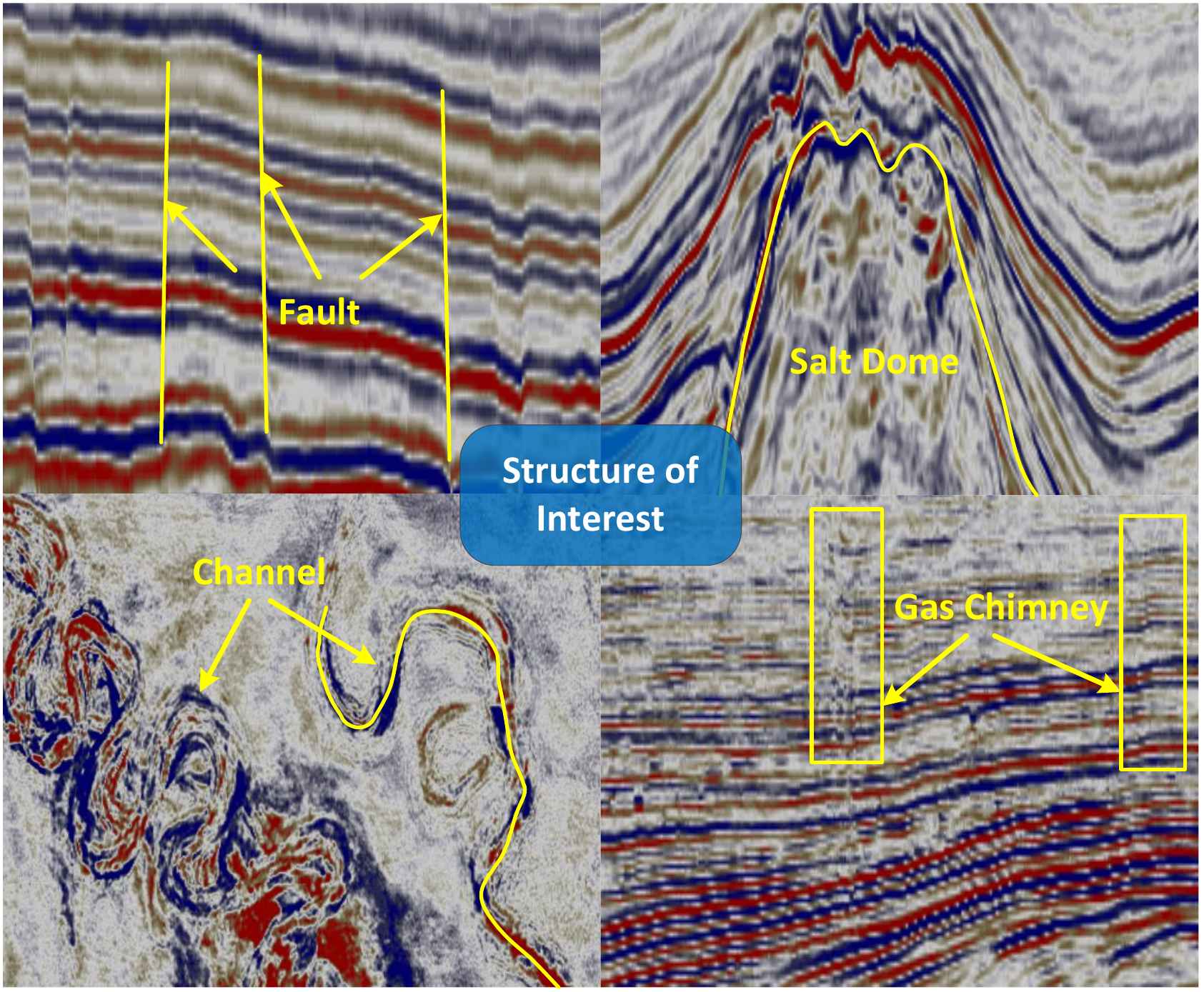}
	\caption{Examples of four types of subsurface structures essential for hydrocarbon exploration.}
	\label{figure:structuresofinterest}
\end{figure}

Subsurface structures are complex because of the massive geologic evolution and deformation over millions of years. A migrated seismic volume, therefore, can contain multitudes of geologic structures such as horizons, unconformities, faults, salt domes, channels, and gas chimneys. Horizons, represented as seismic reflections, are the dominant geologic structures apparent in a seismic volume while the other structures can often be recognized as discontinuities or edges of seismic reflections. We will focus the discussion in this paper on the interpretation of faults, salt domes, channels, and gas chimneys. Examples are shown in Fig. \ref{figure:structuresofinterest}. All these structures are of great geological implications for hydrocarbon migration and accumulation as discussed below. 

\noindent \textbf{Faults}: A fault is defined as a lineament or planar surface across which apparent relative displacement occurs in the rocks layers. The movement of impermeable rocks and sediments along the fault surface creates membranes that hinder the migration of hydrocarbons from source rocks and create structural hydrocarbon traps. Because of the lateral changes in texture across a fault, interpreting such structures can be treated as an edge detection problem that is common in digital image processing. However, different from edges in natural images, faults do not always display explicit edges, and the visual appearance is typically noisy. 

\noindent \textbf{Salt Domes}: A salt dome is defined as a dome-shaped structure formed by the evaporation of a large mass of salt in sedimentary rocks. Salt domes are impermeable structures that prevent the migration of hydrocarbons and provide entrapment for oil and gas reservoirs. Chaotic reflections are often observed in a salt dome in a form of a distinct texture; thereby interpreting such structures can be treated as a texture segmentation problem in digital image processing. Nevertheless, in salt domes, the boundaries are not explicit due to the underlying physics as well as noise and low-resolution data. 

\noindent \textbf{Channels}: A channel is defined as a remnant of an inactive river or a stream channel that has been either filled or buried by younger sediment. Channels, as well as the associated depositional features such as overbanks, lobs, and fans, are important targets of seismic interpretation because river/stream flows often carry sands of high porosity and permeability that are superior traps for hydrocarbon accumulation. Such rock difference gives channels a unique, distinct texture, similar to salt domes; meanwhile considering its meandering spatial distribution, interpreting such structures can be treated as a curve boundary detection and tracking in digital image processing.

\noindent \textbf{Gas Chimneys}: A gas chimney is defined as the leakage of gas from a poorly sealed hydrocarbon accumulation in the subsurface. Gas chimneys are often used as a robust hydrocarbon indicator, which implies present or previous vertical migration of hydrocarbons or fluids containing hydrocarbons between different geologic sequences. In a seismic profile, a gas chimney is visible as a vertical zone of poor data quality or push-downs. Therefore, a gas chimney is often featured with a distinct texture. Correspondingly, similar to salt domes, interpreting such structures can be treated as a texture segmentation problem in digital image processing.

\subsection{Common Geophysical Datasets}
High-quality seismic data are essential for accurate subsurface interpretation with broad applications to both industrial applications (e.g., oil \& gas exploration) and environmental studies (e.g., earthquake monitoring and $CO_2$ storage). Driven by the oil and gas industry, 3D seismic data have been widely collected around the world in the past few decades, particularly in the areas with high hydrocarbon potential, such as the North Sea, the Middle East, and the Gulf of Mexico. Such data were often confidential during the initial stage of reservoir exploration and production but are then released for public use when the production goal has been achieved. Here, we list in Table~\ref{table_datasets} seven seismic datasets that are commonly used for developing and testing interpretation algorithms. The table includes the geographical location of the surveys, the dominant subsurface structures, the data size, as well as the spatial dimensions. The first six datasets are available in the typical SEG-Y format, which can be loaded directly using any geophysical exploration software (e.g., OpendTect and Petrel\textsuperscript{\textregistered}) or popular programming packages such as MATLAB\textsuperscript{\textregistered} and Python. As to the LANDMASS dataset, it is available in the MATLAB\textsuperscript{\textregistered} MAT-File format and provides a large number of image patches generated from the F3 block to facilitate structure-based seismic interpretation and machine learning studies. 
All links to these datasets and other resources can be found at \url{https://ghassanalregib.com/datasets-and-resources-for-computational-seismic-interpretation/}. 
\begin{table}[!htbp]
\centering
\caption{List of commonly-used datasets for interpretation\cite{CeGPCSI2017}.}
\label{table_datasets}
\resizebox{0.98\textwidth}{!}{
\begin{tabular}{|l|l|c|c|l|}
\hline
\textbf{Dataset \& Location} & \textbf{Dominant Structures}  & \textbf{Size} &\textbf{Domain} & \textbf{Dimensions}\\ 
\hline
\hline
F3 block in the Netherland North Sea & \begin{tabular}[c]{@{}l@{}}Salt domes ($\sim$1500 ms); Faults ($\sim$1200 ms);\\   Gas chimneys ($\sim$500 ms)\end{tabular}      & 494 MB        & Time            & Inline: 651, Crossline: 951, Samples/trace: 463                         \\
\hline
Stratton field in south Texas, USA     & Channels ($\sim$845 ms); Faults ($\sim$2000 ms)               & 122 MB        & Time            & Inline: 100, Crossline: 200, Samples/trace: 151              \\
\hline
Teapot Dome in Wyoming, USA        & Faults ($\sim$5500 ft)                 & 421 MB        & Depth           & Inline: 345, Crossline: 188, Samples/trace: 1601                              \\
\hline
Great South Basin in New Zealand     & Faults ($\sim$2000 ms)                     & 38.1 GB       & Time            & Inline: 1241, Crossline: 2780, Samples/trace: 751                                        \\
\hline
Waka Basin in New Zealand        & Channels ($\sim$1800 ms)                     & 23.2 GB       & Time            & Inline: 801, Crossline: 5756, Samples/trace: 1500                                                                                                                           \\
\hline
SEAM synthetic dataset          & Salt domes ($\sim$2000 ms)                  & 4.2 GB        & Time \& Depth            & Inline: 1169, Crossline: 1002, Samples/trace: 851                              \\
\hline
\hline
\multirow{2}{*}{LANDMASS dataset from F3 block}           & Horizons, Chaotic reflections,   & \multirow{2}{*}{0.95 GB}   & \multirow{2}{*}{Time}    & Horizon patches: 9385, Chaotic patches: 5140, \\
& Faults and Salt domes & & & Fault patches: 1251, Salt dome patches: 1891\\
\hline
\end{tabular}
}
\end{table}

\section{Subsurface Event Detection and Tracking}
\label{sec:detNtrk}

\begin{figure}[!htbp]
\centering
\begin{tikzpicture}[scale=0.6, every node/.style={scale=0.6, font=\large}]
\path[mindmap,
      concept color=violet,
      text=white,
      level 1/.append style={level distance = 4.5cm, sibling angle = 90, text width = ,minimum size = 2.5cm},
      level 2/.append style={level distance = 4cm, text width =, minimum size = 3cm},
      level 3/.append style={level distance = 4cm, sibling angle = 50, text width = , minimum size = 2cm},
      ]

 node[concept] {Subsurface Structure Interpretation}
 [clockwise from = 120]
 child[concept color = JungleGreen] {
   node[concept](Faults) {Faults}
   [clockwise from = 120]
   child { 
        node[concept](FaultDetection) {Detection} 
        [clockwise from = 160]
        child[concept color = orange]{node[concept, align = center]{Ant \\ Tracking}}
        child[concept color = orange, level distance = 5cm]{node[concept, align = center, scale = 1.1]{Hough \\ Transform~\cite{wang2017fault}\cite{wang20143DHough}}}
        child[concept color = orange, level distance = 5cm]{node[concept, scale = 1.4](coherence){Coherence~\cite{Alaudah2106eage}}}   
   }
   child { 
        node[concept](FaultTracking) {Tracking} 
        [clockwise from = 50]
        child[concept color = orange, level distance = 6cm]{node[concept, scale = 1.3]{Interpolation~\cite{wang2017fault}}}
   }
 }
 child[concept color= Maroon] {
   node[concept] {Salt Dome}
   [clockwise from = 100]
   child { 
        node[concept] {Tracking} 
        [clockwise from = 40]
        child[concept color = orange, level distance = 6cm]{node[concept, align = center, scale = 0.9]{Tensor-based \\ Subspace Learning~\cite{wang2015salt}}}
   }
   child { 
        node[concept](SaltDetection) {Detection} 
        [clockwise from = 0]
        child[concept color = orange, level distance = 5cm] {node[concept, align = center, scale = 1.3]{Edge \\ Detection~\cite{asjad2015}}}
        child[concept color = orange] {node[concept, align = center](GoT){Gradient of \\ Texture (GoT)~\cite{Shafiq2017_GoT_Intr}}}
        child[concept color = orange] {node[concept](GLCM){GLCM}}
   }
 }
 child[concept color= Magenta, level distance = 5cm] {node[concept](Gas){Gas Chimney}}
 child[concept color= RoyalBlue, level distance = 6cm, grow = -150]{node[concept](Channel){Channel} };
\node[annotation, left, font=\small, text width = 4.5cm] at (FaultDetection.south west)
  {Fault Detection:\\
  $\bullet$ Discontinuity along horizons\\
  $\bullet$ Curved shapes
  };
\node[annotation, right, font=\small, text width = 4.5cm, ] at (FaultTracking.north east)
  {Tracking: strong correlations\\
  between neighboring sections
  };
\node[annotation, above right = 0.5cm, font=\small, text width = 5cm, ] at (SaltDetection.north west)
  {Salt Dome Detection: Texture\\
  changes near salt dome boundaries
  };
\node[annotation, right, font=\small, text width = 4.6cm] at (Gas.east)
  {Gas Chimney Interpretation:\\
   texture changes around borders
  };
\node[annotation, left, font=\small, text width = 5.5cm] at (Channel.north)
  {Channel Interpretation: Discontinuity\\
  with meandering shapes
  };
\begin{scope}[on background layer]
\path (coherence) to[circle connection bar switch color=from (orange) to (RoyalBlue)] (Channel);
\path (coherence) to[circle connection bar switch color=from (orange) to (Maroon)] (SaltDetection);
\path (GLCM) to[circle connection bar switch color=from (orange) to (RoyalBlue)] (Channel);
\path (GoT) to[circle connection bar switch color=from (orange) to (Magenta)] (Gas);
\end{scope}
\end{tikzpicture}
\caption{Subsurface structures and their corresponding interpretation methods.}
\label{fig:diagramDT}
\end{figure}
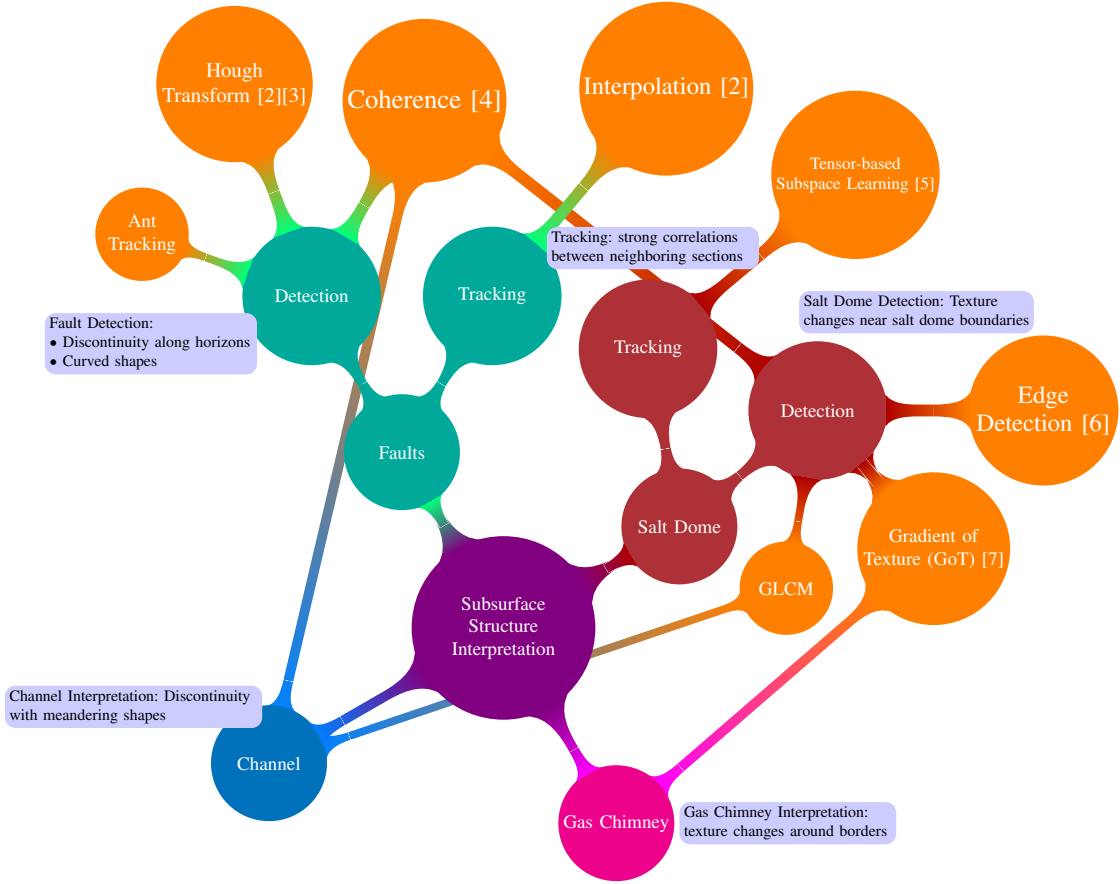

A successful reservoir exploration requires a reliable identification of indicative subsurface structures as introduced in the previous section. Consequently, the majority of existing interpretation algorithms focus on detecting such structures or events in geophysical terms. We have recently proposed an additional task that tracks such structures by creating a semi-automated interpretation workflow. In this section, we review these methods for each of the four key structures, respectively. A depiction of all such algorithms is shown in Fig.~\ref{fig:diagramDT}. 
 
\subsection{Fault Detection}
\label{ssec:faultDetection}
In a 2D seismic section, faults indicate displacements along fractures, as observed in Fig.~\ref{figure:structuresofinterest}. 
Because of the geological constraints that are associated with the formation process, faults have two specific features. One is the geological feature, which is the discontinuity along horizons. The other is the geometric feature, i.e., line-like or curved shapes in 2D seismic sections, which appear as curved surfaces in a 3D seismic volume. Computational fault detection methods are commonly developed based on these two features. The discontinuity of faults can be characterized by several seismic attributes such as entropy~\cite{cohen2006detection}, curvature~\cite{roberts2001curvature}, and coherence~\cite{marfurt19983, gersztenkron1999coherence, bakker2002image}. Among them, coherence is the most popular one for highlighting faults. Marfurt \textit{et al}.~\cite{marfurt19983} calculated coherence by comparing the dissimilarity of local regions on the two sides of a fault. Later,  
Gersztenkorn and Marfurt~\cite{gersztenkron1999coherence} proposed the eigenstructure-based coherence, which analyzes the eigenstructure of covariance matrices of windowed seismic traces. Recently, a generalized-tensor-based coherence (GTC) attribute~\cite{Alaudah2106eage} has been proposed.
The GTC of every voxel in a seismic volume is calculated within a local analysis cube with the size of $I_1\times I_2\times I_3$, which can be represented by a third-order tensor, denoted by $\mathbfcal{A}\in\mathbb{R}^{I_1\times I_2\times I_3}$. 
Unfolding $\mathbfcal{A}$ along three modes generates matrix $\mathbf{A}_{(i)}\in\mathbb{R}^{I_{i}\times \prod_{j\neq i} I_j}$, $i=1,2,3$. Using an eigenstructure analysis on the covariance matrix of unfolding matrix $\mathbf{A}_{(i)}$, the coherence attributes of three modes are obtained as follows: 
\begin{equation}
\label{equ:GTC}
E_{c}^{(i)} = \lambda_1^{(i)}/Tr(\mathbf{C}_i) =  \lambda_1^{(i)}/Tr\left[ (\mathbf{A}_{(i)}-\mathbf{1}_i\otimes \boldsymbol{\mu}_i)^T(\mathbf{A}_{(i)}-\mathbf{1}_i\otimes 
\boldsymbol{\mu}_i)\right]\\,
\end{equation}
where $E_{c}^{(i)}$ is the coherence attribute of mode-$i$, and $\lambda_{1}^{(i)}$ represents the largest eigenvalue of covariance matrix $\mathbf{C}_i$. In addition, $\mathbf{1}_i$ is a matrix of ones with a size of $1\times \prod_{j\neq i}I_j$,  $\boldsymbol{\mu}$ defines the mean of all columns in $\mathbf{A}_{(i)}$, and $\otimes$ represents the Kronecker product. If we take $E_{c}^{(i)}$, $i=1,2,3$, as the R, G, and B channels of a color image, we arrive at the GTC attribute. In contrast to~\cite{gersztenkron1999coherence}, GTC enhances the details of seismic data and allows a better fine-tuning and more flexibility to interpreters. In some cases, depending on one seismic attribute does not produce accurate fault delineation. Therefore, enhancement operations such as non-linear mapping and structure-oriented filtering~\cite{fehmers2003fast} may be applied to increase the contrast between faults and surrounding structures. In addition, combining various attributes~\cite{di2017fault} is another common practice in fault interpretation. 

\begin{figure}[h]
	\centering
	\includegraphics[width =17cm]{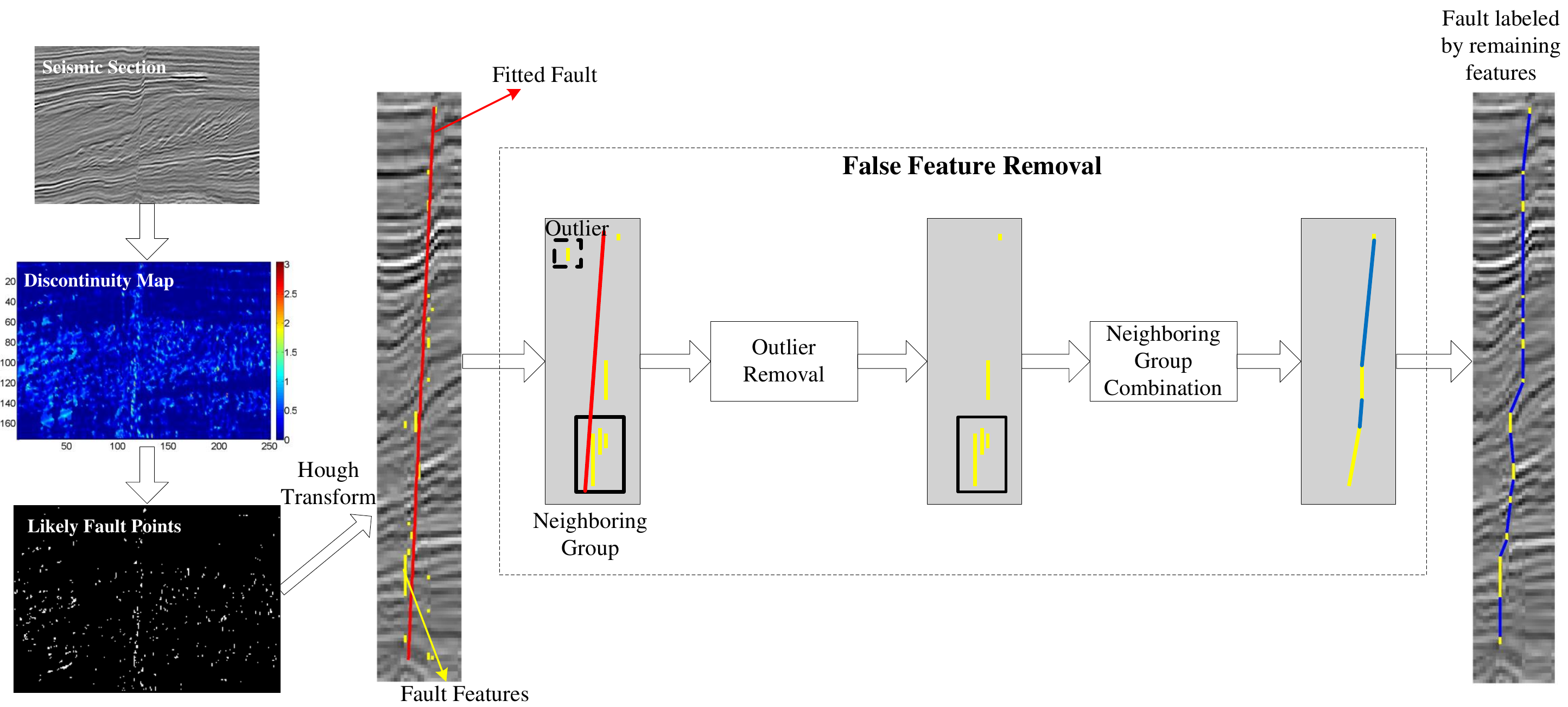}
	\caption{The diagram of the proposed fault detection method, which utilizes the Hough transform to extract fault features and remove false ones using geological constraints. }
	\label{fig:falseRemoval}
\end{figure}

In the attribute space, likely fault regions are commonly specified using a hard threshold or selected from local maxima. In the work of Hale~\cite{hale2013methods}, faults points with the largest fault likelihoods are selected and connected to construct meshes of quadrilaterals, which are further used to form fault surfaces. However, likely fault regions that may inevitably involve noisy structures are not able to accurately reveal the details of faults. To solve this problem, interpreters utilize the geometric feature of faults. Because of the continuous curved shapes of faults, ant tracking or ant colony optimization algorithm~\cite{pedersen2002automatic} has become a popular method for fault detection. Recently, Wu and Hale~\cite{wu20163d} proposed to use a simple linked data structure that includes the fault likelihood, dip, and strike to construct complete fault surfaces. In addition, Hough transform, a mapping procedure between attribute volumes and the parameter space, has been widely used to detect lines, circles, planes, and other parametric shapes. Therefore, Wang and AlRegib~\cite{wang2017fault} proposes to delineate faults using line-like features extracted by the Hough transform. Fig.~\ref{fig:falseRemoval} illustrates the diagram of the proposed method, in which authors first highlight likely fault points by applying a hard threshold on the discontinuity map and then extract line-like yellow fault features using the Hough transform. Because of the limitation of the Hough transform, it is inevitable that detected results contain some false features that violate certain geological constraints. On the basis of their appearance, false features are classified into two types: \textit{outliers}, which are isolated from the others, and \textit{neighboring groups}, which contain similar features located nearby. In Fig.~\ref{fig:falseRemoval}, the red line is fitted from all midpoints of fault features. Therefore, features with larger distances to the fitted fault are treated as outliers and discarded. In contrast, features in a neighboring group will be merged into one fault feature. 
After false feature removal, the remaining features are connected using blue lines to implement the delineation of faults. The same concept of Hough transform has also been applied to extract fault planes from discontinuity volumes~\cite{wang20143DHough}. 

\subsection{Salt Dome Detection}
\label{ssec:saltDetection}
To detect salt domes, over the past a few decades, researchers have proposed various seismic attributes and automated workflows based on edge detection, texture, machine learning, graph theory, active contours, and other techniques. Aqrawi \emph{et al.} \cite{aqrawi2011detecting} proposed a Sobel edge-detection-based approach that delineates salt domes within the Gulf of Mexico dataset. Amin and Deriche~\cite{asjad2015} proposed an approach that highlights small variations in seismic data by detecting edges not only along the x, y, and z directions but also slanted at $45$ and $-45$ degrees. The 2D Sobel filters along $45^{\circ}$ and $-45^{\circ}$ are shown as follows: 
\begin{equation}
\label{equ:sobel3D}
S_{45^{\circ}}=
\begin{bmatrix}
-2 & -1 & 0 \\
-1 & 0 & 1 \\
0 & 1 & 2
\end{bmatrix}
,
S_{-45^{\circ}}=
\begin{bmatrix}
0 & -1 & -2 \\
1 & 0 & -1 \\
2 & 1 & 0\\
\end{bmatrix}.
\end{equation}
Berthelot \emph{et al.} \cite{berthelot2013texture} proposed a Bayesian classification approach for detecting salt bodies using a combination of seismic attributes such as dip, similarity, frequency-based attributes, and attributes based on the gray-level co-occurrence matrix (GLCM). A workflow based on seismic attributes and a machine-learning algorithm (i.e., extremely random trees ensemble) that automatically detects salt domes from the SEAM dataset is presented in \cite{ml_paperPablo_15}. Amin and Deriche \cite{asjadcodebook2016} proposed a supervised codebook-based learning model for salt-dome detection using texture-based attributes. Qi \emph{et al.} Di \emph{et al.} \cite{Di_EAGE2017} proposed an interpreter-assisted approach based on the k-means clustering of multiple seismic attributes that highlights salt-dome boundaries in the F3 block. Shafiq \emph{et al.} \cite{Shafiq_ICASSP2016} proposed a seismic attribute called \emph{SalSi}, which highlights salient areas of a seismic image by comparing local spectral features based on the 3D Fast Fourier Transform (FFT). Chopra and Marfurt \cite{ChopraMarfurtSD_2016} proposed a seismic disorder attribute to assess randomness and the SNR of data to delineate seismic structures such as faults and salt domes. Wu \cite{Wu_Geophysics_2016} proposed methods to compute salt likelihoods highlighting salt boundaries, extract oriented salt samples on ridges of salt likelihoods, and construct salt boundaries with the salt samples by solving a screened Poison surface reconstruction problem.

\begin{figure}[h]
\centering
\sbox{\measurebox}{%
  \begin{minipage}[b]{.42\textwidth}
  \subfloat
    []
    {\label{fig:figA}\includegraphics[width=8cm]{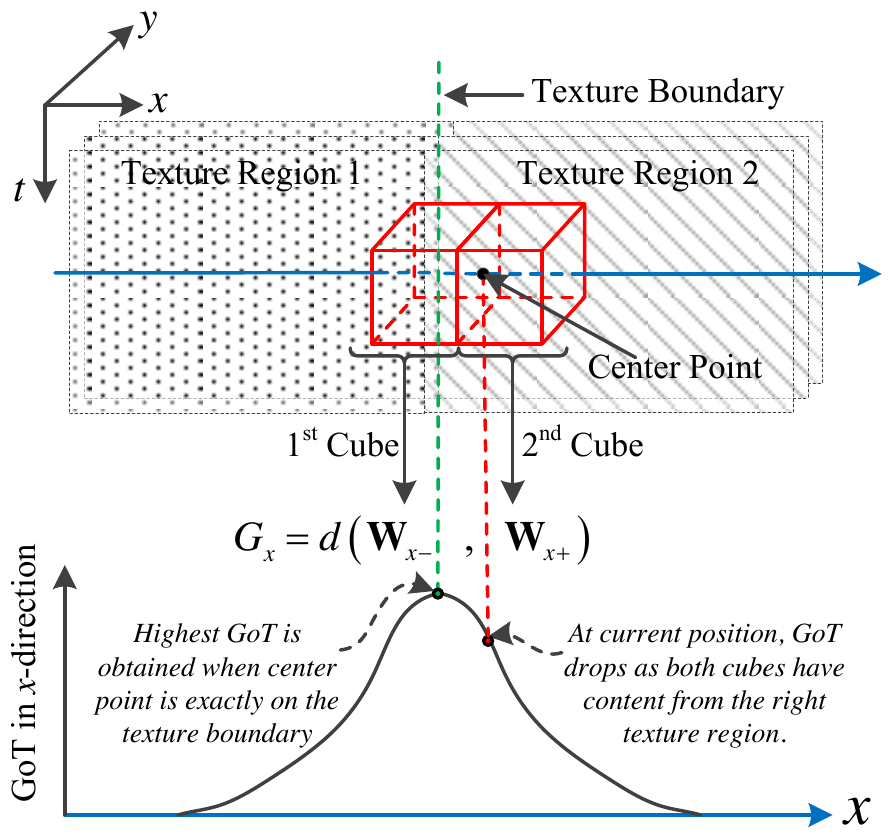}}
  \end{minipage}}
\usebox{\measurebox}\qquad
\hspace{1cm}
\begin{minipage}[b][\ht\measurebox]{.42\textwidth}
\centering
\subfloat
  []
  {\label{fig:figB}\includegraphics[width=6cm]{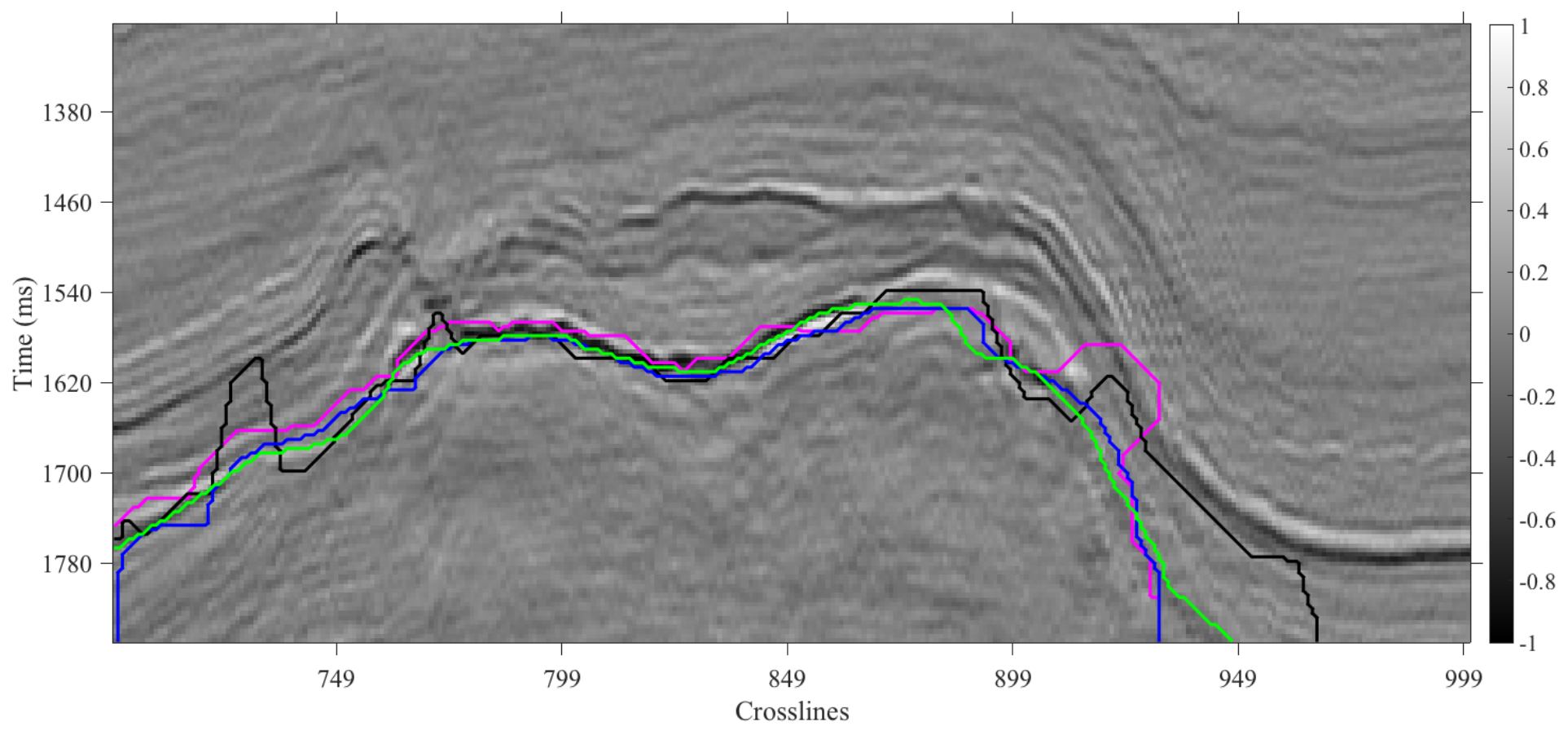}}
\vspace{-.45cm}
\subfloat
  []
  {\label{fig:figC}\includegraphics[width=6cm]{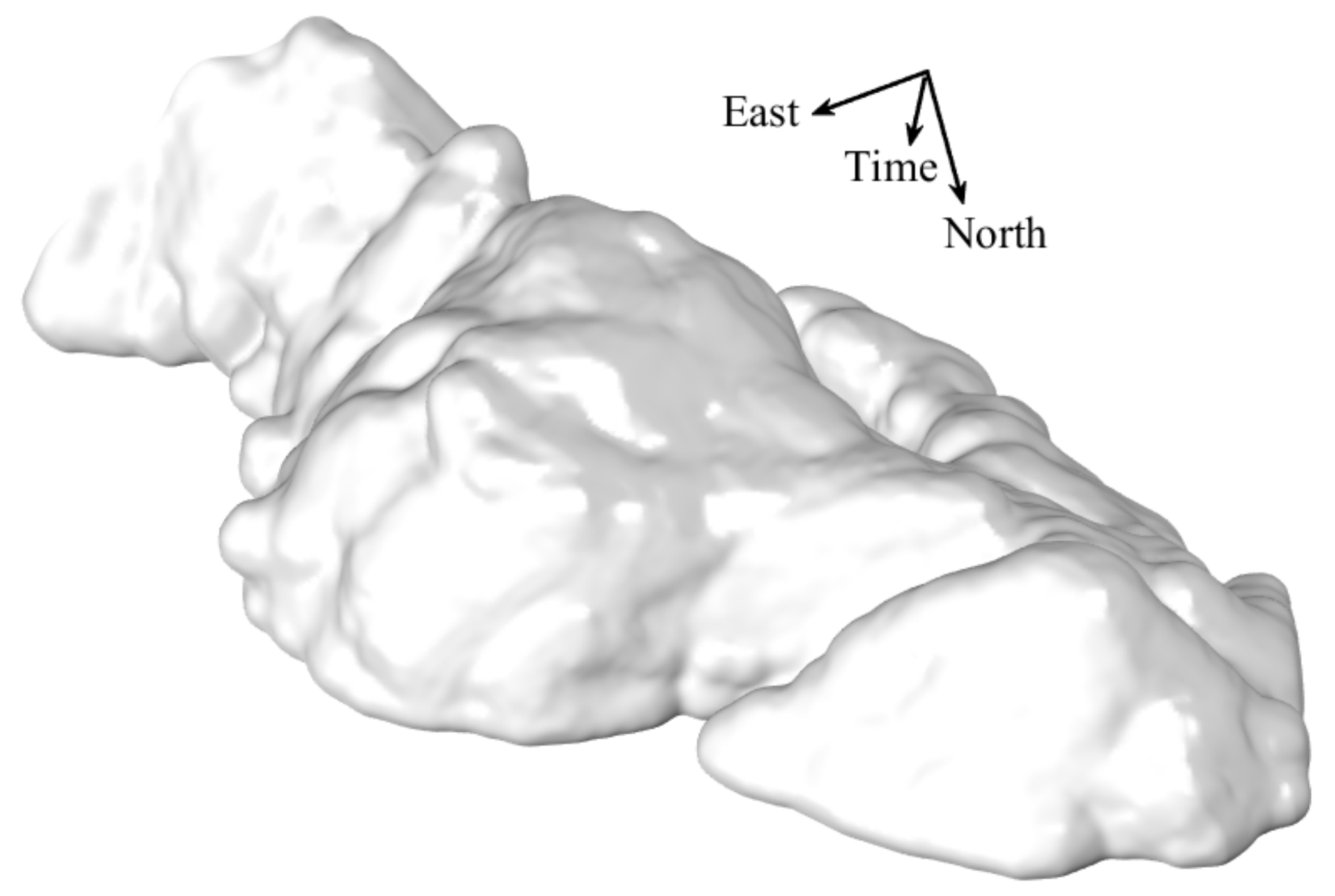}}
\end{minipage}
\caption{(a) Computing GoT along the crossline, i.e., $x$-direction. (b) Salt-dome boundary detected by various delineation algorithms. Magenta: \cite{aqrawi2011detecting}, Black: \cite{asjadcodebook2016}, Blue: \cite{Shafiq2017_GoT_Intr}, Green: Reference salt-dome boundary. (c) A 3D salt body detected from the F3 block in the North Sea using 3D-GoT~\cite{Shafiq2017_GoT_Intr}.}
\label{fig:GoTFigure}
\end{figure}

With most of the above algorithms, the underlying assumption is that a salt dome body can be recognized by comparing neighboring planes. Thus, capturing the textures of such planes has been a very effective methodology to delineate salt dome bodies. To show the effectiveness of texture-based approaches for detection of salt domes, the brief details of a seismic attribute, three-dimensional gradient of textures (3D-GoT) by Shafiq \emph{et al.} \cite{Shafiq2017_GoT_Intr}, which describes the texture dissimilarity between neighboring cubes around each voxel in a seismic volume across time (or depth), crossline, and inline directions is presented in this section. Fig.~\ref{fig:GoTFigure}(a) illustrates synthetic seismic images in which a green dashed vertical line separates two textured regions depicted in dotted and striped lines, respectively. To evaluate the GoT in the $x$-direction (i.e., the crossline), as the center point and its two neighboring cubes move along the blue line, texture dissimilarity function $d(\cdot)$ yields a GoT profile shown at the bottom of Fig.~\ref{fig:GoTFigure}(a). Theoretically, the highest GoT value corresponds to the highest dissimilarity and is obtained when the center point falls exactly on the texture boundary. Similarly, GoT is also calculated along $t$- (time or depth) and $y$- (inline) directions. To improve the delineation efficiency and robustness, GoT employs a multi-scale gradient expressed as follows: 
\begin{align}
\label{equ:MultiGoT}
\mathbf{G}[t,x,y]= \left( \sum\limits_{i\in\{t,x,y\}} \left(\sum\limits_{n=1}^{N}\omega_n\cdot d\left(\mathbf{W}_{i-}^{n}, \mathbf{W}_{i+}^{n}\right)\right)^2 \right)^\frac{1}{2},
\end{align}
where $\mathbf{W}_{i-}^n$ and $\mathbf{W}_{i+}^n$ denote the neighboring cubes, $n$ represents the edge length of cubes, and $\omega_n$ represents the weight associated with each cube size. To compute dissimilarity between cubes, authors use a \emph{perceptual} dissimilarity measure based on error magnitude spectrum chaos, which is not only computationally less expensive and performs better than \emph{non-perceptual} dissimilarity measures, but also highlights texture variations in the most effective manner. The perceptual dissimilarity measure is calculated as follows: 
\begin{align}
d(\cdot) =  E \left( \left| \mathcal{K}   \oasterisk   \left| \mathcal{K} \oasterisk \left|\mathbf{W}_{i-}^{n} - \mathbf{W}_{i+}^{n}\right|  \right|  \right| \right), \quad i\in\{t,x,y\},
\end{align}
where $\oasterisk$ represents the tensor product, $\mathcal{K}$ is the Kronecker matrix defined as $\mathcal{K} = D_t \otimes D_x \otimes D_y$, and $D_t$, $D_x$, and $D_y$ are DFT transformation matrices. The output of various salt dome delineation algorithms on a typical seismic inline and a salt body detected using 3D-GoT \cite{Shafiq2017_GoT_Intr} from the F3 block in the North Sea are shown in Fig.~\ref{fig:GoTFigure}(b-c), respectively.

\subsection{Fault and Salt-dome Tracking}
\label{ssec:faultSaltTracking}
\begin{figure}[h]
	\centering
	\includegraphics[width = 16cm]{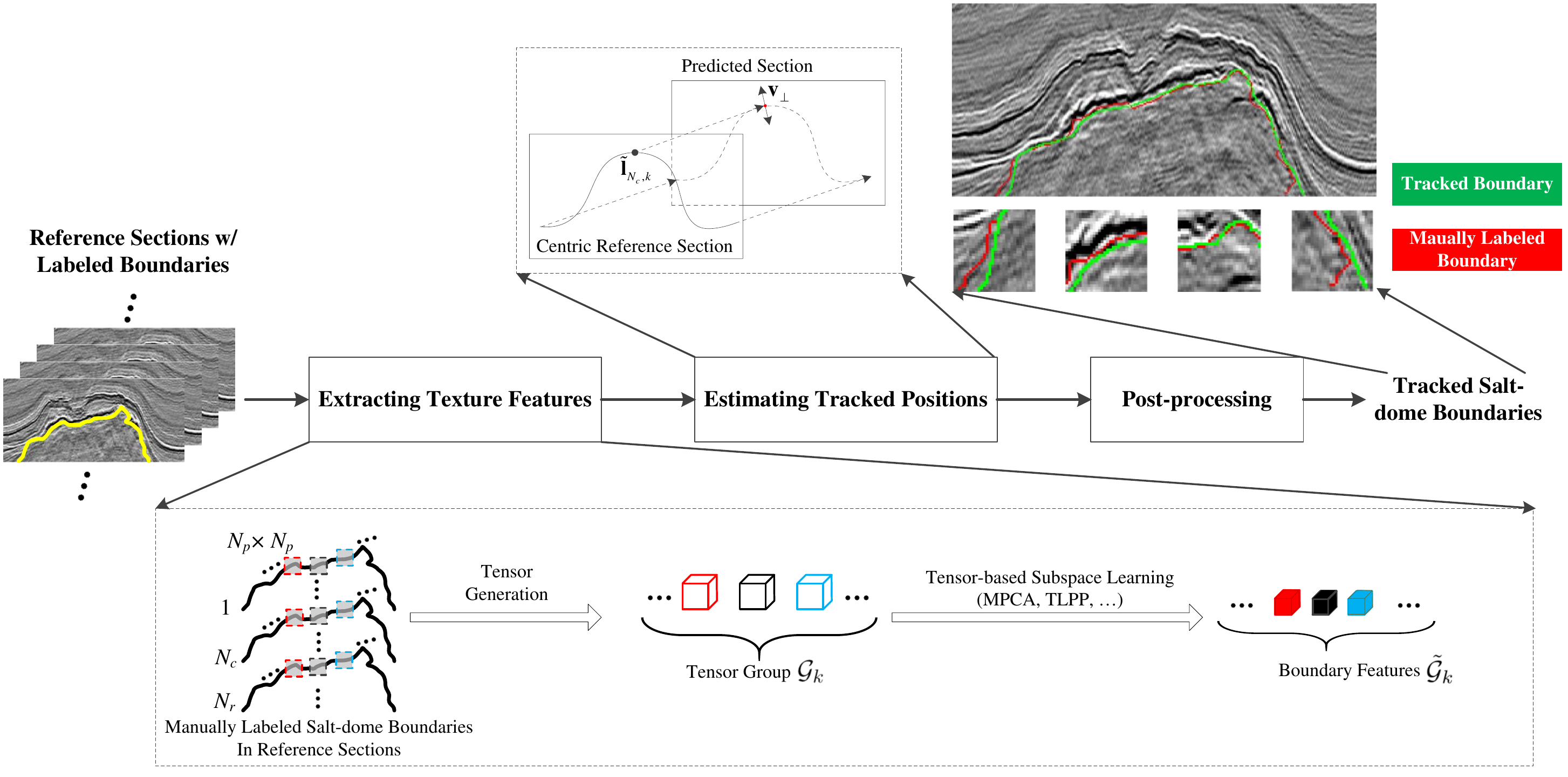}
	\caption{The diagram of salt-dome tracking, in which features around salt-dome boundaries in reference sections are extracted by tensor-based subspace learning.}
	\label{fig:subspace}
\end{figure}

The methods introduced in the previous two sections focus mainly on the detection, or delineation, of faults and salt domes in 2D sections. To investigate the geological structures of faults and salt domes, interpreters need to repeatedly apply these methods on each section of a seismic volume. However, for a seismic volume with a large size, the repeated detection in every section may impair interpretation efficiency. Because of the slow formation processes of subsurface structures, neighboring sections commonly have strong correlations. In recent years, fault and salt-dome tracking methods have been proposed to utilize correlations between sections to improve interpretation efficiency. \cite{wang2017fault} borrows the concept of motion vectors in video coding and grouped seismic sections into reference and predicted sections. Faults in the predicted sections can be labeled by detection results in reference sections. Berthelot \textit{et al.}~\cite{berthelot20123d} detected the salt-dome boundary in one time-section using the method in~\cite{berthelot2013texture} and tracked the boundary through adjacent sections by minimizing a defined energy function that maintains boundaries' curvature and smoothness. Similarly,~\cite{Shafiq_EAGE2016_AC} takes advantage of active contour to track salt-dome boundaries in neighboring seismic sections. More recently, Wang \textit{et al.}~\cite{wang2015salt} have proposed a salt-dome tracking method that extracts the features of salt-dome boundaries in reference sections using tensor-based subspace learning and delineates tracked boundaries by finding points in predicted sections, which are the most similar to reference ones. Fig.~\ref{fig:subspace} illustrates the diagram of feature extraction process using tensor-based subspace learning adopted for salt-dome tracking. Authors manually label the boundaries of salt domes in $N_r$ reference sections, in which $N_c=\lceil N_r/2\rceil$ is the central one. For each point in the labeled boundary of $N_c$, its corresponding points in neighboring $(N_r-1)$ reference boundaries are identified. Since every point corresponds to a patch with the size of $N_p\times N_p$, stacking corresponding patches constructs third-order texture tensors from reference sections, denoted $\left\{\mathbfcal{A}_{k}\in\mathbb{R}^{N_p\times N_p\times N_r}, k=1,2,\cdots,K_{N_c}\right\}$, where $K_{N_c}$ represents the number of points in the $N_c$-th reference boundary. Because of strong correlations between neighbors, every point in the $N_c$-th reference boundary corresponds to a tensor group, denoted $\mathbfcal{G}_k=\left\{\mathbfcal{A}_{k-N_s},\cdots,\mathbfcal{A}_{k},\cdots,\mathbfcal{A}_{k+N_s}\right\}$, which contains $2N_s+1$ tensors. Using multilinear dimensionality reduction methods such as multilinear principle component analysis (MPCA) and tensor-based linearity projection preserving (TLPP), one can obtain transformation matrices $\mathbf{U}_{k}^{(1)}$, $\mathbf{U}_{k}^{(2)}$, and $\mathbf{U}_{k}^{(3)}$ and map tensor group $\mathbfcal{G}_k$ to its subspace as follows: 
\begin{equation}
\label{equ:tensorElement}
\begin{aligned}
\tilde{\mathbfcal{A}}_{m}=\mathbfcal{A}_m\times_1\mathbf{U}_k^{(1)}\times_2&\mathbf{U}_k^{(2)}\times_3\mathbf{U}_k^{(3)}\mbox{, }\\
&m\in\left\{k-N_s,\cdots,k+N_s\right\}, 
\end{aligned}
\end{equation}
where $\times_i$, $i=1,2,3$, denotes the $i$-mode product of a tensor by a matrix, and tensor $\tilde{\mathbfcal{A}}_{m}$ with less dimensions in each mode contains the features of reference boundaries. These extracted features can help identify tracked boundary points in predicted sections. The local and global comparison of the tracked boundary (green) with the manually labeled one (red) shows the high similarity and accuracy of salt-dome tracking within seismic volumes.

\subsection{Channel Detection}
The variety of rocks that are deposited in a channel system change the acquired seismic signals, and thereby a channel system can be detected by applying the common edge detection methods on horizontal slices or well-picked horizon surfaces. For example, the coherence attribute \cite{bahorich19953}\cite{wu2017directional} was first proposed for identifying the morphology of a channel system, and then a suite of new attributes and algorithms have been developed for enhancing both detection accuracy and noise robustness. However, while successfully highlighting the channel boundaries, such discontinuity analysis fails to depict the rock properties within the channel system, as no such information is considered by the detection operators. To overcome this limitation, the GLCM analysis was introduced from the field of image processing and used for seismic facies analysis~\cite{gao2011latest}. Eichkitz \emph{et al.} \cite{eichkitz2013calculation} presented a suite of algorithms for generating ten different GLCM attributes from a seismic cube. Di and Gao \cite{di2017nonlinear} made it possible to decompose a channel system by integrating the logistic/exponent gray-level transformations with the GLCM attributes. Machine learning-based facies analysis has gained popularity, particularly for a deposition system with complexities, and automatically classified all subtle structures, including the meandering channel, overbanks, fans, and lobs. The use of computational classification began soon after the development of seismic attributes in the 1970s with the work by Justice \emph{et al.} \cite{justice1985multidimensional}. Barnes and Laughlin \cite{barnes2002investigation} reviewed several unsupervised classification techniques, including k-means, fuzzy clustering, and self-organizing maps (SOMs) and emphasized the importance of seismic attributes over the classifiers. Wallet \emph{et al.} \cite{wallet2009latent} developed the generative topographic mapping for unsupervised waveform classification. Song \emph{et al.} \cite{song2017multi} combined multi-linear subspace learning with the SOMs for improved seismic facies analysis in the presence of noise. A comprehensive study of both supervised and unsupervised facies analysis can be found in Zhao \emph{et al.} \cite{zhao2015comparison}. With these depositional features in a channel system well differentiated by either seismic attributes or facies analysis, they could then be easily extracted as separate geo-bodies by seeded tracking. However, reliable differentiation of various depositional features (e.g., overbank, delta, and levee) remains challenging, due to the proximity and overlying distribution in space between each other and more importantly their similar reflection patterns in 3D seismic data. Such a goal can be achieved by analyzing the seismic images at a smaller scale to capture the subtle differences between various features.

\subsection{Gas-chimney Detection}
In seismic sections, a gas chimney is visible as vertical zones of poor data quality, chaotic reflections, or push-downs. Therefore, it can be detected using attributes similar to those used for salt-dome detection, such as  the coherence \cite{bahorich19953} and the GoT \cite{Shafiq2017_GoT_Intr}. The major difference of gas chimneys is the sparse distribution in a seismic volume, and thereby manual identification and interpretation of them is labor intensive, and the conventional tracking tools described in section \ref{ssec:faultSaltTracking} may also fail in detecting the chimneys that are isolated from each other. However, the machine learning-based approach offers an efficient solution to such limitation. For example, Heggland \emph{et al.} \cite{meldahl1999chimney} combined a set of seismic attributes and the multilayer perception (MLP) to create a chimney cube for a semi-automatic detection, which has been applied to multiple datasets, such as the F3 block \cite{hashemi2008gas} and the Taranaki basin in New Zealand \cite{singh2016interpretation}. In recent years, researchers also have tried more advanced machine learning algorithms. For example, Xiong \emph{et al.} \cite{xiong2014adaboost} applied adaptive boosting (AdaBoost) to the design of the optimal learning algorithm for identifying gas chimneys, which generated more reliable results than the k-nearest neighbor method. Xu \emph{et al.} \cite{xu2017multi} implemented the sparse autoencoder for gas chimney detection, and the accuracy is greatly improved compared to the traditional MLP algorithm. The sparsity of the spatial distribution of gas chimneys in a seismic dataset adds the difficulty of reliable gas-chimney detection in two ways. First, it limits the amount of training data, and thereby the performance of supervised learning may be affected. Weakly-supervised and unsupervised methods would be more applicable. Second, it increases the sensitivity to seismic noise. Reflection pattern-based learning could help improve the noise robustness of gas chimney detection, such as the chaotic labelling described in section \ref{sec:label_class} given the chaotic reflections in a gas chimney.

\section{Subsurface Labeling and Classification}
\label{sec:label_class}

In the previous section, we provided a review of seismic interpretation for subsurface event detection and tracking. Although the techniques have significantly reduced the time and effort required for manual interpretation, there is at least one aspect involved that is still done manually. Namely, the manual process of extracting sub-volumes from a given data volume based on their dominant subsurface structure, so that detection or tracking can be performed on the extracted data. In this section, we discuss a framework we recently developed to address this issue. With this framework, we attempt to eliminate the aforementioned bottleneck and streamline the interpretation process by building on the recent advances in semantic segmentation and scene labeling.

\begin{figure*}[h]
\begin{center}
\includegraphics[width=\linewidth]{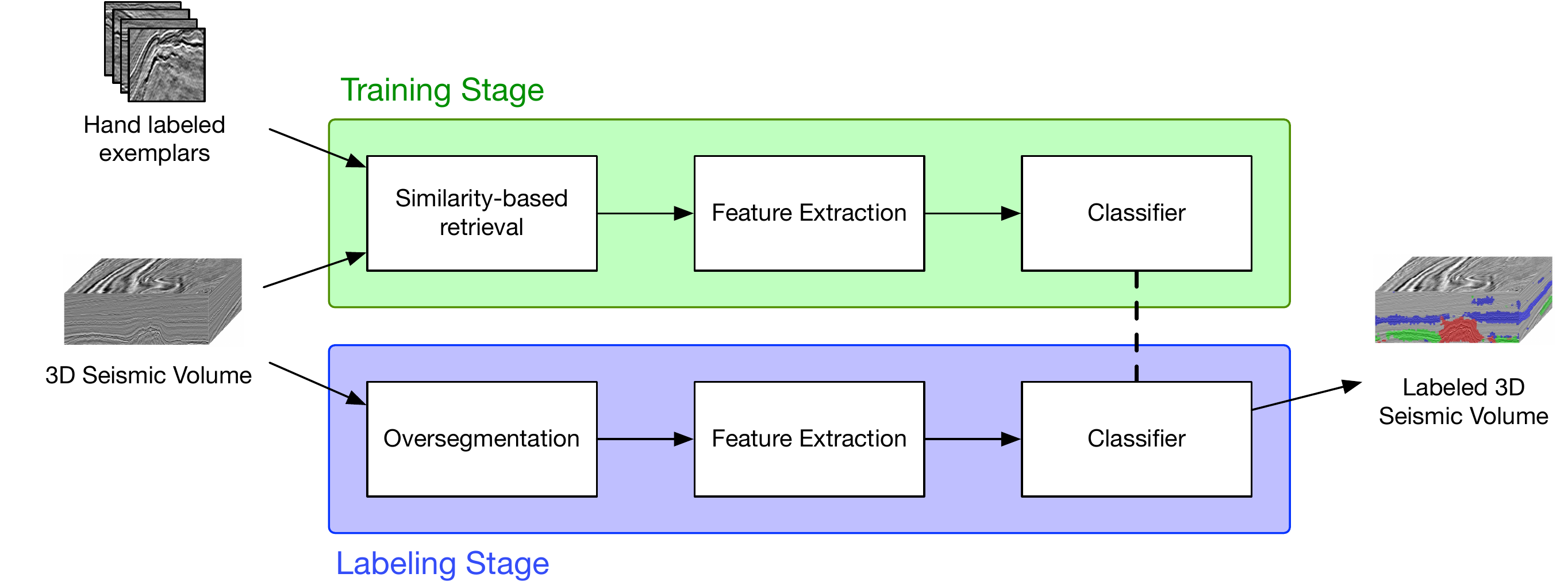}
\end{center}
\caption
{A block diagram illustrating the weakly-supervised seismic volume labeling approach described in section \ref{sec:label_class}.}
\label{fig:workflow3}
\end{figure*}

\subsection{General Framework}\label{generalFramework}

Seismic volume labeling is the process of classifying each voxel in a given seismic volume into one of many predefined structures. This process can help classify entire seismic volumes into regions of interest that contain specific subsurface structures. These regions can then be extracted, and various interpretation algorithms can be applied to these regions for more refined results.

While a variety of influential works have been proposed in the area of semantic segmentation \cite{Zhu2016,FCN,Bansal2016,deeplab,chen2014,Farabet2013}, seismic data presents challenges that cannot be immediately solved by the existing methods. First, unlike natural images where edges between objects are well-defined, edges between subsurface structures in seismic data are either not well-defined or are characterized by a change in overall texture rather than a sudden change in amplitude. Second, unlike natural images, seismic data is grayscale, and thus color features cannot be used to distinguish various structures. Furthermore, there is a severe lack of both labeled seismic data for training and well-established benchmarks for testing various learning-based approaches. This is partly due to the intellectual property concerns in the oil and gas industry. Also, because of the lack of ground truth and the subjective nature of manual interpretation, it is often difficult for different geophysicists to agree on the same interpretation for a given volume. 

Naturally, machine learning techniques are well-suited approaches for seismic volume labeling. However, the lack of labeled training data poses a significant challenge. To tackle this problem, we have generated the LANDMASS \cite{LANDMASS} dataset that contains more than 17,000 seismic images extracted from the Netherlands North Sea F3 block \cite{F3_data}. The images are grouped into four classes based on the subsurface structure they contain, namely, \texttt{horizons}, \texttt{chaotic horizons}, \texttt{faults}, and \texttt{salt domes}. Although the dataset contains these specific structures, the framework for seismic labeling discussed below can be extended to other seismic structures as well. 

In~\cite{ICIP2016_Yazeed}, Alaudah and AlRegib proposed using reference exemplars and seismic image retrieval to label seismic volumes in a weakly-supervised fashion. Firstly, given a few hand-selected exemplar images, $\mathbf{X} = [\mathbf{x}_1, \mathbf{x}_2, \cdots,
\mathbf{x}_{N_e}]$, that contain subsurface structures belonging to the different classes of interest, an augmented dataset, $\tilde{\mathbf{X}}$, is generated in an unsupervised fashion using similarity-based retrieval. This is done to obtain enough data to train a supervised machine learning model, as is further explained in subsection \ref{similarityRetrieval}. Then, various features or attributes are extracted from these images to train a classifier, which is detailed in subsection \ref{textureAttributes}. Subsection \ref{oversegmentation} describes seismic volume segmentation, and its use to enforce the local spatial correlation of the labels and improve the computational efficiency of the approach. Then finally, subsections \ref{volumeLabeling} and \ref{pixelAnnotation} describe various methods to obtain the final labeling of the seismic volume. The overall framework is outlined in Fig.~\ref{fig:workflow3}.

\subsection{Building Blocks}
In the remaining part of this section, we describe the major components of this weakly-supervised approach for seismic volume labeling.  

\subsubsection{Texture Attributes for Seismic Labeling}\label{textureAttributes} 

Seismic images are often well-characterized by texture features, or texture attributes, mainly because they are textural in nature. A few classical texture attributes were explored for traditional tasks such as salt dome detection, but they need to be further examined in the context of seismic labeling. In addition, there are a great number of advanced texture features developed in recent years for image analysis. They are also potential candidates for the labeling task. 

When applied to image processing problems, it is usually desirable for a texture attribute to possess properties such as illumination-, rotation-, and scale-invariance for better robustness. However, with seismic data, this is not always the case. For example, a vertical slice of a seismic volume (or a seismic section) is characterized by strong directionality, with horizons typically extending in the horizontal direction and faults in the vertical direction. In such cases, being rotation-invariant is no longer a critical requirement for the attributes. Another important difference between a seismic image and a natural texture image is that some subsurface structures (e.g., faults) are of very fine scale along certain dimensions, which is not typical with natural textures. Thus, it is important for a texture attribute to be able to capture such fine details.

Recently, some comparative studies were conducted to examine various texture attributes in the context of seismic volume labeling. In one study, the focus was on a group of spatial attributes from the family of local descriptors, including the local binary pattern (LBP), a few of its typical variants, and the local radius index (LRI) \cite{long2017texture}. These attributes can represent texture patterns with robustness and computational efficiency. For comparison purposes, the study also included two traditional seismic attributes in the spatial domain, i.e., the GLCM and the semblance. According to the study, the local descriptors and the GLCM are all good attributes for labeling seismic volumes. However, each attribute displayed different characterizing capabilities for different subsurface structures. Thus, they should be selected accordingly if there is a preference for certain structures to be labeled with more reliability. 

In a separate study \cite{alfarraj2017texture}, multiresolution attributes in the frequency domain were examined for seismic volume labeling, including the discrete wavelet transform and its nonsubsampled version, Gabor filters, the steerable pyramid, the contourlet transform and its nonsubsampled counterpart, and the curvelet transform. Effective singular values are extracted from each transformed subband and then concatenated into a feature vector. The major conclusion from the study is that the directional transforms perform better than the non-directional ones, with the curvelet transform being the best.

\subsubsection{Similarity-based Retrieval}\label{similarityRetrieval} 
The attributes described in the previous section are used in the feature extraction step of the workflow depicted in Fig.~\ref{fig:workflow3}. Also, these features can be used to devise similarity measures specialized for seismic images (e.g., \cite{STSIM2013,ICIP2015Yazeed,MMSP2016Motaz}) to be used in the similarity-based retrieval to generate the augmented data, $\tilde{\mathbf{X}}$. Note that texture-based image similarity metrics are different from the generic fidelity and distance measures (also known as quality metrics, such as Peak Signal-to-Noise Ratio (PSNR) and Mean Square Error (MSE)), in that they capture the content of an image rather than assuming a pixel-to-pixel correspondence. 

One particularly useful similarity measure was proposed in \cite{MMSP2016Motaz} for content-based image retrieval (CBIR), especially for images that are highly textural such as seismic images. The proposed measure uses the singular values of the curvelet transform of an image to form a feature vector. Then, the similarity of any two images is computed as the Czekanowski similarity between their corresponding feature vectors as $
\textsc{Similarity} (\mathbf{x}_1,\mathbf{x}_2) = 1-\frac{\|\bar{\mathbf{v}}_1-\mathbf{v}_2\|_1}{\|\mathbf{v}_1+\mathbf{v}_2\|_1}$,
where $\mathbf{x}_1$, $\mathbf{x}_2$ are the images and $\mathbf{v}_1$, $\mathbf{v}_2$ are their corresponding feature vectors. Since the singular values of real images are non-negative by definition, the value of this similarity measure is in the range of $[0,1]$, where a value of $1$ indicating the two images are identical. 

With a proper similarity measure in hand and an exemplar image for each subsurface structure of interest, the training data is formed by retrieving images that are most similar to the exemplar image. The training dataset is then used to train a machine learning model as will be detailed in section \ref{volumeLabeling}. 

\subsubsection{Over-segmentation of Seismic Volumes} \label{oversegmentation} 

An important part of the workflow in Fig.~\ref{fig:workflow3} is the seismic volume over-segmentation. Segmentation algorithms, like normalized cuts \cite{ncuts}, are often used in computational seismic interpretation to extract  subsurface structures \cite{SaltSegmentation_Halpert_14, hale_segmentation_1, Lomask_normalized_cuts_2}. However, over-segmentation here is used as a preprocessing step to enforce local spatial correlation by grouping voxels in the volume that are similar and close to each other. Here, the lack of clearly-defined edges between subsurface structures is inconsequential, and instead, each small segment is classified based on its texture content. This step also significantly reduces the computational cost of the labeling, since each segment will be classified once, and the resulting label will then be propagated to all the voxels within that segment.

Graph-based over-segmentation techniques (such as graph cuts \cite{gcuts} and normalized cuts \cite{ncuts}) or gradient ascent based approaches (such as the watershed algorithm \cite{watershed} and turbo pixels \cite{turbo}) are computationally expensive, and therefore not suitable for large seismic volumes. Simpler, and more efficient techniques such as SLIC superpixels \cite{slic} are more appropriate for this application. Since SLIC clusters pixels in the CIELAB color space, while seismic images are grayscale, it is possible to obtain a good superpixel over-segmentation by replacing the A and B channels with the $x-$ and $y-$ gradients of the seismic image \cite{ICIP2016_Yazeed}.  

Once the over-segmentation is performed, features are extracted from each segment, or from a group of segments centered around a reference segment. These features are later used to classify the voxels within these segments into one of the various subsurface structures that are of interest to the interpreter. 

\subsubsection{Labeling of Seismic Volumes}\label{volumeLabeling} 

Once the augmented data $\tilde{\mathbf{X}}$ is created using the methods described in section \ref{similarityRetrieval}, texture features can then be extracted from these images to train various machine learning models for labeling the seismic volume. These models can be support vector machines (SVMs), convolutional neural networks (CNNs), random forests, or any other machine learning model suitable for this task. Once these models are trained, we can then extract texture attributes from each segment in the over-segmented seismic volume, and classify them. The resulting label is then propagated to all voxels within the segment. 

\begin{figure*}[ht]
\begin{centering}
\subfloat{\includegraphics[width=0.48\textwidth]{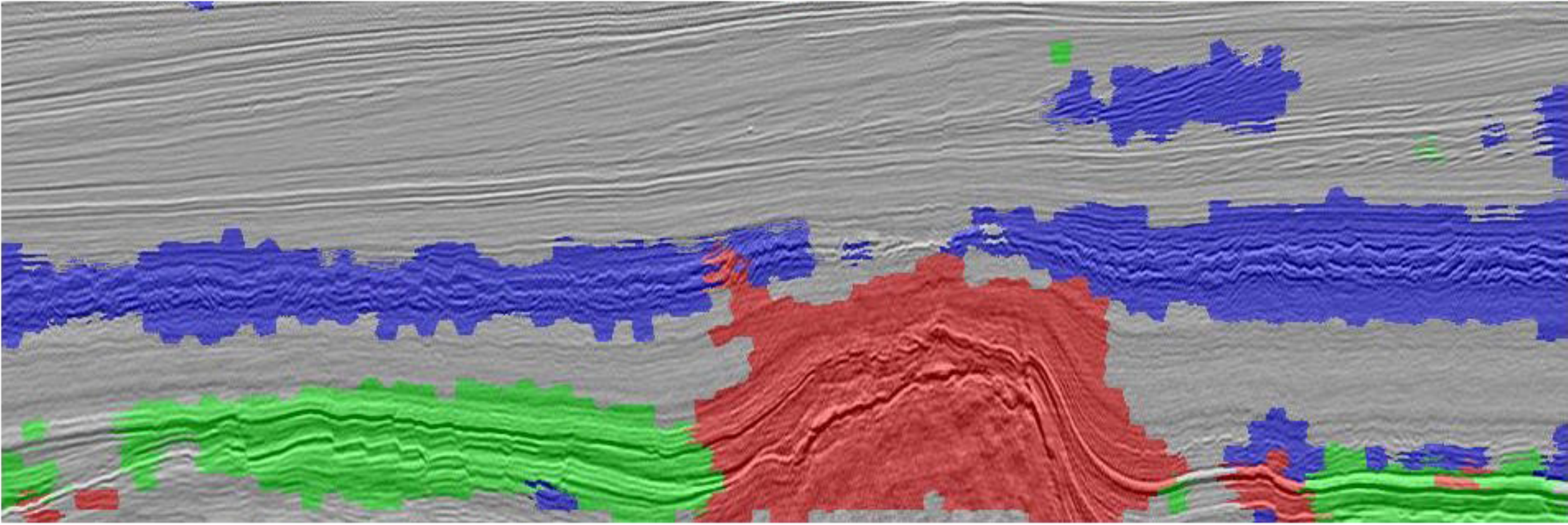}}
\hfill 
\subfloat{\includegraphics[width=0.48\textwidth]{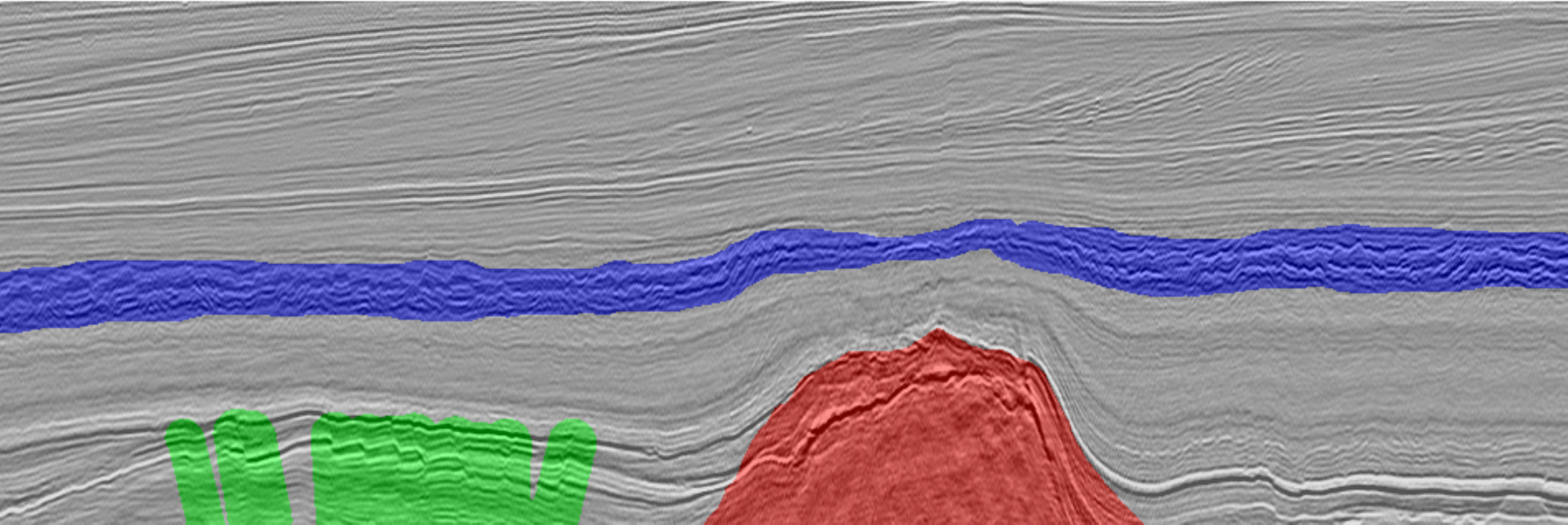}}
\caption{On the left, a weakly-supervised labeled seismic section from the Netherlands North Sea F3 block database. For reference, the manually labeled seismic section is shown on the right. The \texttt{chaotic} class is in blue, \texttt{faults} is in green, \texttt{salt dome} is in red.}
\label{fig:labeling}
\end{centering}
\end{figure*}

The ultimate goal of seismic volume labeling is to efficiently and accurately classify entire seismic volumes based on their subsurface structures such as shown in Fig.~\ref{fig:labeling}. There will naturally be a trade-off between the accuracy and the efficiency of any algorithm. However, it is better to obtain a rough labeling quickly, rather than an accurate labeling that is time-consuming. Other specialized techniques would later be used for the refined detection and tracking of the subsurface structures.

\subsubsection{Weakly-supervised Pixel-level Annotation}\label{pixelAnnotation}

\begin{figure}[!htbp]
	\centering
	\includegraphics[width = 6 in]{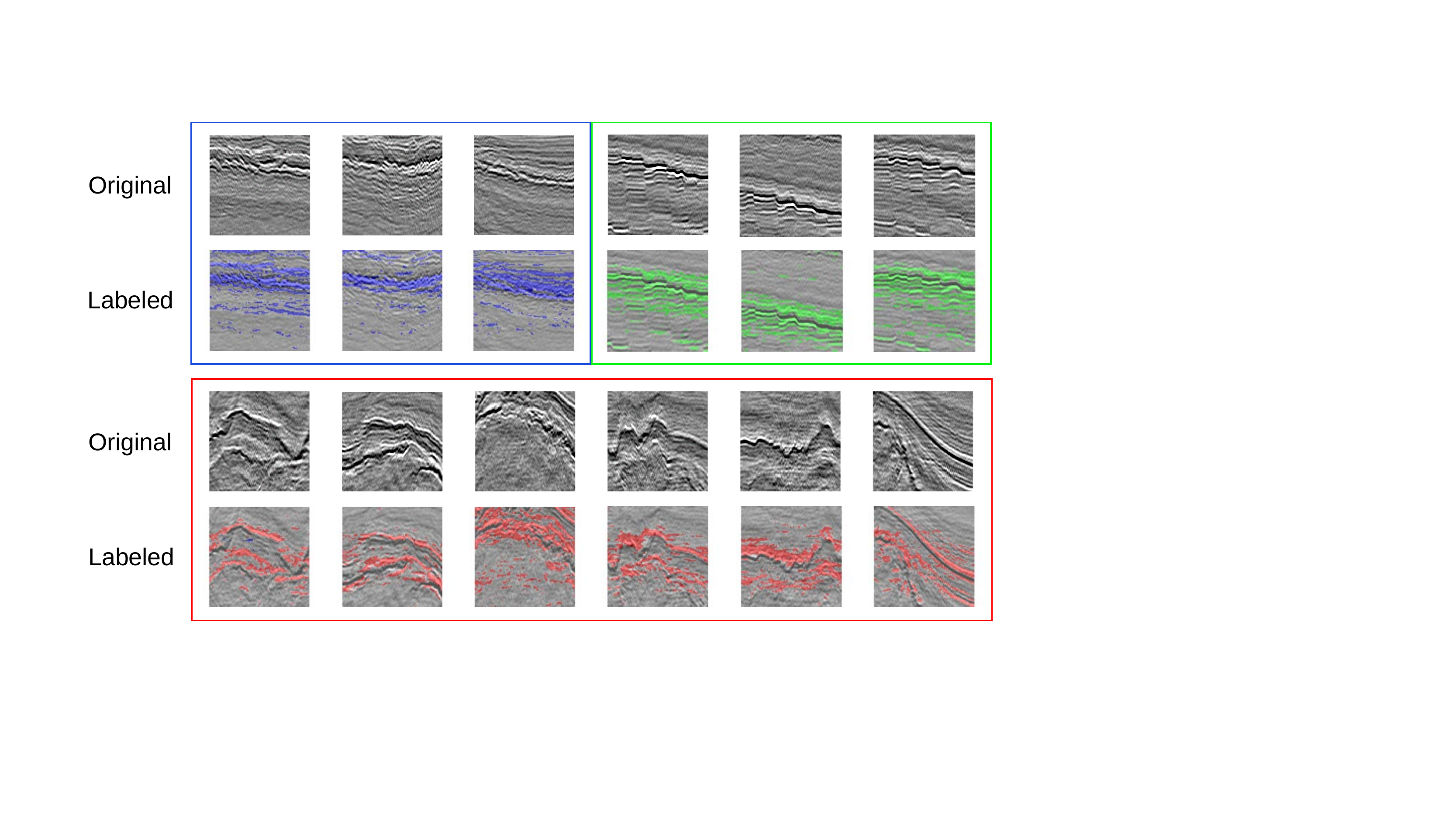}
    \caption{Results of the weakly-supervised pixel-level annotation approach in \cite{SEG2017_WSL_Yazeed} for various subsurface structures. The first row shows images containing \texttt{chaotic} and \texttt{fault} structures and the second row shows their corresponding pixel-level labels in blue and green respectively. The third and fourth rows show various images that contain \texttt{salt dome} bodies or boundaries, and their corresponding pixel-level labels in red.}
	\label{fig:WSL}
\end{figure}

In section \ref{similarityRetrieval} we have shown that similarity-based retrieval can be used to extract thousands of images that have the same subsurface structure to generate our augmented  training data $\tilde{\mathbf{X}}$. The retrieved images are then assigned the same image-level label as the query image. While these image-level annotations can be used to train machine learning models, recent work on weakly-supervised labeling such as that proposed in \cite{SEG2017_WSL_Yazeed} show that these image-level labels can be mapped into pixel-level labels that can be much more useful in training powerful fully-supervised deep learning models such as fully convolutional neural nets (FCN) \cite{FCN}. Fig.~\ref{fig:WSL} shows several examples of this mapping for various subsurface seismic structures. 

The approach in \cite{SEG2017_WSL_Yazeed} is based on non-negative matrix factorization (NMF) \cite{Lee2001}, and is summarized in Algorithm \ref{alg:wsl}. NMF is a commonly used matrix factorization technique that is closely related to many unsupervised machine learning techniques such as $k$-means and spectral clustering \cite{Ding2005}. NMF  decomposes a non-negative matrix $\tilde{\mathbf{X}} \in \mathbb{R}_+^{ N_p \times N_s}$ into the product of two lower-rank matrices $\mathbf{W} \in \mathbb{R}_+^{N_p \times N_f}$, and $\mathbf{H} \in \mathbb{R}_+^{N_f \times N_s}$ such that both $\mathbf{W}$ and $\mathbf{H}$ are non-negative.  In other words,
\begin{equation}
\tilde{\mathbf{X}} \approx \mathbf{W}\mathbf{H} ~~~~\mathrm{s.t.} \mathbf{W},\mathbf{H} \geq 0.  
\end{equation}

\noindent Here, $N_f$ is the number of components (or the \emph{rank}) of $\tilde{\mathbf{X}}$. In our work, $\tilde{\mathbf{X}}$ represents the augmented data matrix from section \ref{similarityRetrieval} where each column is a single seismic image in vector form. The data matrix  $\tilde{\mathbf{X}}$ has $N_s$ such images, each of which is a vector of length $N_p$. NMF factorizes this data matrix into two non-negative matrices; a basis matrix $\mathbf{W}$ and a coefficient matrix $\mathbf{H}$. 

The regular NMF problem does not have an analytical solution, and is typically formulated as the following non-convex optimization problem: 
\begin{equation}\label{NMF}
\underset{\mathbf{W},\mathbf{H}}{\arg\min}  || \mathbf{X} - \mathbf{W}\mathbf{H} ||_F^2  ~~~~\textrm{s.t.}  \mathbf{W}\mathbf{H} \geq 0,
\end{equation}
\noindent where, $||\cdot||_F$ is the Frobenius norm. Lee \textit{et al.} \cite{Lee2000} showed that NMF can be used to learn a ``parts-based" representation, where each feature would represent a localized ``part" of the data. In practice, this is rarely achieved using the formulation in equation \ref{NMF}. To remedy this, the feature matrix $\mathbf{W}$ is initialized using $k$-means applied on each class in the data matrix $\tilde{\mathbf{X}}$ separately. This initialization simplifies the feature learning, and makes $\mathbf{W}$ robust to miss-labeled images in $\tilde{\mathbf{X}}$. Then a sparsity constraint is imposed on these initial features using the following sparsity measure:  

\begin{equation}
\rho(\mathbf{w}) = \frac{\sqrt{N_p} - ||\mathbf{w}||_1 / ||\mathbf{w}||_2}{\sqrt{N_p} - 1},
\end{equation}

\noindent where $||\cdot||_1$ and $||\cdot||_2$ are the $l_1$ and $l_2$ norms respectively, and $\rho(\cdot)$ indicates the sparsity of a vector.  To enforce this constraint, we follow the algorithm proposed by \cite{Hoyer2004}. Additionally, to make sure that each feature $\mathbf{w}_i$ in the matrix $\mathbf{W}$ represents a single class only, we impose an orthogonality constraint on the coefficients matrix $\mathbf{H}$. We also add two regularization terms on $\mathbf{W}$ and $\mathbf{H}$ to avoid overfitting. The problem then becomes: 

\begin{equation}
\begin{aligned}\label{SONMF}
\underset{\mathbf{W},\mathbf{H}}{\arg\min}  || \tilde{\mathbf{X}} - \mathbf{W}\mathbf{H} ||_F^2  +& \lambda_1 ||\mathbf{W}||_F^2 + \lambda_2||\mathbf{H}||_F^2  +  \gamma_1||\mathbf{H}\mathbf{H}^T - \mathbf{B}||_F^2 \\
&\textrm{s.t.}  \mathbf{W}, \mathbf{H} \geq 0 ~~\mathrm{and}~~  \rho(\mathbf{w}_i) = \rho_w,
\end{aligned}
\end{equation}

\noindent where matrix $\mathbf{B} \in \mathbb{R}^{N_f \times N_f}$ contains random positive real numbers, and $\lambda_1$, $\lambda_2$, $\gamma_1$, and $\rho_w$ are constants. To solve the problem in equation \ref{SONMF}, the following multiplicative update rules for $\mathbf{W}$, and $\mathbf{H}$ are derived, where

\begin{equation}\label{WMUR}
\mathbf{W}^{t+1} = \frac{ \mathbf{W}^{t} \odot(\tilde{\mathbf{X}}{\mathbf{H}^t}^T)_{ij}}{(\mathbf{W}^t\mathbf{H}^t{\mathbf{H}^{t}}^T +\lambda_1\mathbf{W}^{t})_{ij}}
\end{equation}
\begin{equation}\label{HMUR}
\mathrm{and} ~ \mathbf{H}^{t+1} =  \frac{\mathbf{H}^t \odot({\mathbf{W}^{t+1}}^T \tilde{\mathbf{X}} + \gamma_1 (\mathbf{B} + \mathbf{B}^T) \mathbf{H}^{t})_{ij}}{ {(\mathbf{W}^{t+1}}^T\mathbf{W}^{t+1}\mathbf{H}^{t} + \gamma_1\mathbf{H}^t{\mathbf{H}^{t}}^T\mathbf{H}^t +\lambda_2\mathbf{H}^{t} )_{ij}}.
\end{equation}

\noindent Here, $\odot$ represents element-wise multiplication, and the superscript indicates the iteration number. Once $\mathbf{W}$ and $\mathbf{H}$ are initialized, the multiplicative update rules in equation \ref{WMUR} and \ref{HMUR} are applied successively until both $\mathbf{W}$ and $\mathbf{H}$ converge. 

Once $\mathbf{W}$ and $\mathbf{H}$ have converged, each column of $\mathbf{H}$, $\mathbf{h}_n$, indicates the features used to construct the  $n^{th}$ image. Since every feature in $\mathbf{W}$ should correspond to a single seismic structure, the coefficients of each image can then be mapped into the seismic structures that make up that image. Thus for every image $n \in [1,N_s]$ we can obtain:       

\begin{equation}\label{7}
\mathbf{L}_n = \mathbf{W}^{\mathsf{final}} (\mathbf{Q} \odot (\mathbf{h}_n^{\mathsf{final}} \mathbf{1}_{1 \times N_l} ) )  ~~~~~ \forall n \in [1,N_s],
\end{equation}

\noindent  where $\mathbf{Q} \in \{0,1\}^{N_f \times N_l}$ is a cluster membership matrix such that the element $Q_{ij} = 1$ if the feature $\mathbf{w}_i$ belongs to structure $j$, and $\mathbf{1}_{1\times N_l}$ is a vector of ones of size $1 \times N_l$. The resulting matrix, $\mathbf{L}_n\in \mathbb{R}_+^{N_p \times N_l}$ shows the likelihood of each seismic structure for each pixel in the image. Then, the pixel-level label for each location $i$ in image $n$ corresponds to the seismic structure given by 

\begin{equation}\label{8}
\tilde{\mathbf{y}}_n = \arg \underset{j}{\max}  ~~ \mathbf{L}_{n}(:,j).
\end{equation}

\noindent This method is summarized in Algorithm \ref{alg:wsl}, and sample results are shown in Fig. \ref{fig:WSL}.

\pagebreak

\begin{algorithm}
\SetAlgoLined
\KwIn{Augmented data matrix $\tilde{\mathbf{X}}  \in \mathbb{R}^{N_p \times N_s}$, image-level labels $\mathbf{y} \in \mathbb{R}^{1 \times N_s}$, number of classes $N_l$, and feature sparsity value $\rho_w$.}
\KwOut{Pixel-level labels matrix $\tilde{\mathbf{Y}} \in \mathbb{R}^{N_p \times N_s}$.}

$\mathbf{W}^0 = \mathsf{kMeans}(\tilde{\mathbf{X}}, k)$ \;

$\mathbf{W}^0 = \mathsf{applySparsityConstraint}(\mathbf{W}^0, \rho_w)$ \;

$\mathbf{H}^0 = \mathsf{randomInitialization}()$ \;

$\mathbf{Q} = \mathsf{constructClusterMembershipMatrix}(\mathbf{y}, N_l)$\;

\While{not converged}{
    $\mathbf{W}^{t+1} = \frac{ \mathbf{W}^{t} \odot(\tilde{\mathbf{X}}{\mathbf{H}^t}^T)_{ij}}{(\mathbf{W}^t\mathbf{H}^t{\mathbf{H}^{t}}^T +\lambda_1\mathbf{W}^{t})_{ij}}$ \; 
    
    $\mathbf{H}^{t+1} =  \frac{\mathbf{H}^t \odot({\mathbf{W}^{t+1}}^T \tilde{\mathbf{X}} + \gamma_1 (\mathbf{B} + \mathbf{B}^T) \mathbf{H}^{t})_{ij}}{ {(\mathbf{W}^{t+1}}^T\mathbf{W}^{t+1}\mathbf{H}^{t} + \gamma_1\mathbf{H}^t{\mathbf{H}^{t}}^T\mathbf{H}^t +\lambda_2\mathbf{H}^{t} )_{ij}}$ \;

    $\mathbf{H}^{t+1} =\mathsf{normalizeColumns}(\mathbf{H}^{t+1})$\;
    
    t = t + 1\;
 }
 \For{$n \leftarrow 1$ \KwTo $N_s$}{

   $\mathbf{h}_n^{\mathsf{final}}  =  \mathbf{H}^{\mathsf{final}}(:,n) $\;
   
    $\mathbf{L}_n = \mathbf{W}^{\mathsf{final}} (\mathbf{Q} \odot (\mathbf{h}_n^{\mathsf{final}} \mathbf{1}_{1 \times N_l} ) )   $ \;

    $\tilde{\mathbf{y}}_n = \arg \underset{j}{\max}  ~~ \mathbf{L}_{n}(:,j)$ \;
 }

 $\tilde{\mathbf{Y}} = [\tilde{\mathbf{y}}_1, \tilde{\mathbf{y}}_2, \cdots , \tilde{\mathbf{y}}_{N_s}]$  \;

\caption{Weakly-Supervised Pixel-Level Annotation}
\label{alg:wsl}
\end{algorithm}

\section{Emerging Trends and Open Problems}
\label{sec:emerging}
From the recent advancement in seismic interpretation research as we have reviewed above, we observe two important factors that contribute to the success of this endeavor. First, to address the challenges rooted in the ever-increasing data size and complexity, it becomes very critical to leverage the advanced machine learning techniques, especially those based on deep learning. Second, being a unique type of visual signals, seismic volumes can be interpreted effectively using image analysis algorithms that utilize human visual system (HVS) characteristics and models. In this section, we will briefly discuss the emerging trends and open problems regarding these two aspects.\\

\begin{figure}[ht!]
	\centering
	\includegraphics[width = 4in]{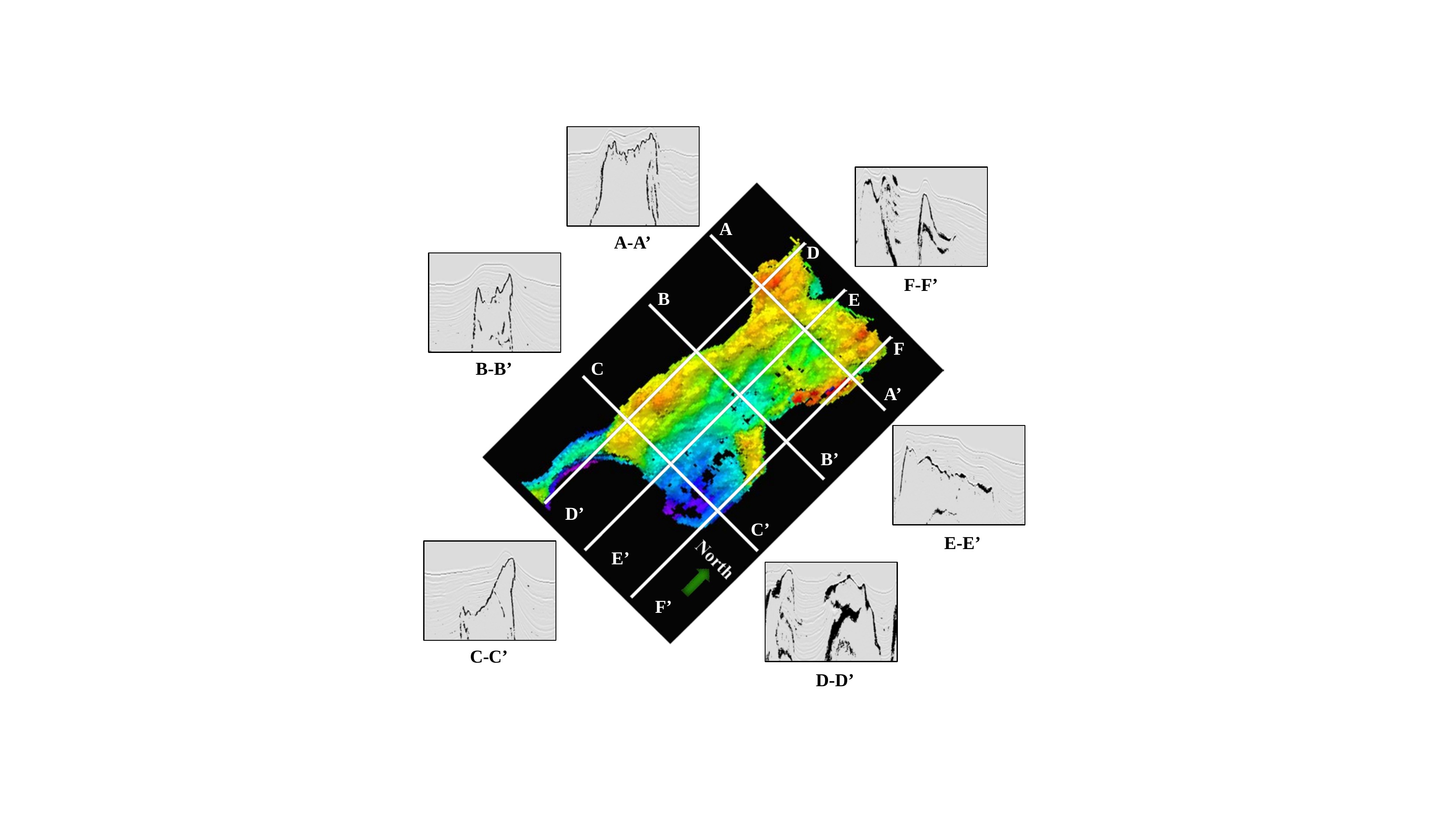}
	\caption{An illustration of implementing CNNs for salt body boundary delineation from the post-stack seismic amplitude. The detected boundaries are clipped to six vertical sections for quality control. Note the good match between the CNN detection and the poststack seismic images}
	\label{figure:cnnsalt}
\end{figure}

\textit{Deep Subsurface Learning}: Deep learning is one of the most powerful learning techniques available nowadays. As a data-driven approach, it utilizes sophisticated neural networks with deep architectures to uncover complex hidden structures and characteristics directly from a large amount of samples. When applied to subsurface data, deep learning will allow geoscientists to make sense out of the massive datasets with many variables while avoiding the human biases. Naturally, seismic interpretation based on deep learning is emerging as a very promising trend. For example, Waldeland and Solbergd \cite{waldeland2017salt} applied convolutional neural networks (CNNs) to classify salt bodies from 3D seismic datasets. Huang \emph{et al.} \cite{huang2017scalable} provided an excellent demonstration of the effectiveness of integrating CNNs and multi-attribute analysis from post-stack amplitude in detecting faults. An illustration of implementing CNNs for salt body boundary delineation is given in Fig. \ref{figure:cnnsalt}. In this case, a good match is observed between the detected boundaries and the original seismic images. Besides the conventional post-stack data used for interpretation, geophysicists are turning their attention to the prestack data. For example, Hami-Eddine \emph{et al.} \cite{hamieddine2017growing} investigated a machine learning approach to optimize the use of both prestack and post-stack seismic data. Araya-Polo \emph{et al.} \cite{araya2017automated} and \cite{lin2017towards} proposed using a deep neural network to learn a mapping relationship between the raw seismic data and the subsurface geology so that the labor-intensive processing stage could be avoided.

To fully explore the potential of deep learning, there are a few open problems that need to be addressed carefully. First, the lack of labeled data is a serious obstacle. Unlike natural image classification problems, public-domain datasets with large sets of labeled samples are rare for seismic interpretation. This severely limits the application of supervised learning techniques. Alternatively, weakly-supervised or unsupervised learning will be more realistic choices in this case. Other techniques such as Generative Adversarial Networks (GANs) and active learning, which will be discussed shortly, can also provide an alternative approach to supervised learning. Second, more advanced network architectures need to be explored for interpretation. Current works mainly use CNNs as the core deep learning structure. To account for the strong correlations along different dimensions in a seismic volume, networks such as the Long Short-Term Memory (LSTM)~\cite{greff2016lstm} and the Neural Turing Machines (NTM)~\cite{graves2014neural} can be incorporated. Third, as a data-driven approach, deep learning relies heavily on the provided data. Thus, a concern with deep learning is about falsely labeled data. Such samples will be misleading in the training of the network. To tackle this problem, various regularization techniques such as dropout can be adopted. If applicable, weakly-supervised or unsupervised learning should be adopted. Finally, seismic imaging and processing that precede the interpretation will inevitably introduce errors or uncertainties. How do these errors propagate through the learning networks? How do they affect the learning performance? Moreover, in general, how do we quantify the uncertainty associated with these learning approaches? These are all issues of practical significance for seismic interpretation, and should be carefully addressed.\\

\begin{figure}[ht]
	\centering
	\includegraphics[width =4 in]{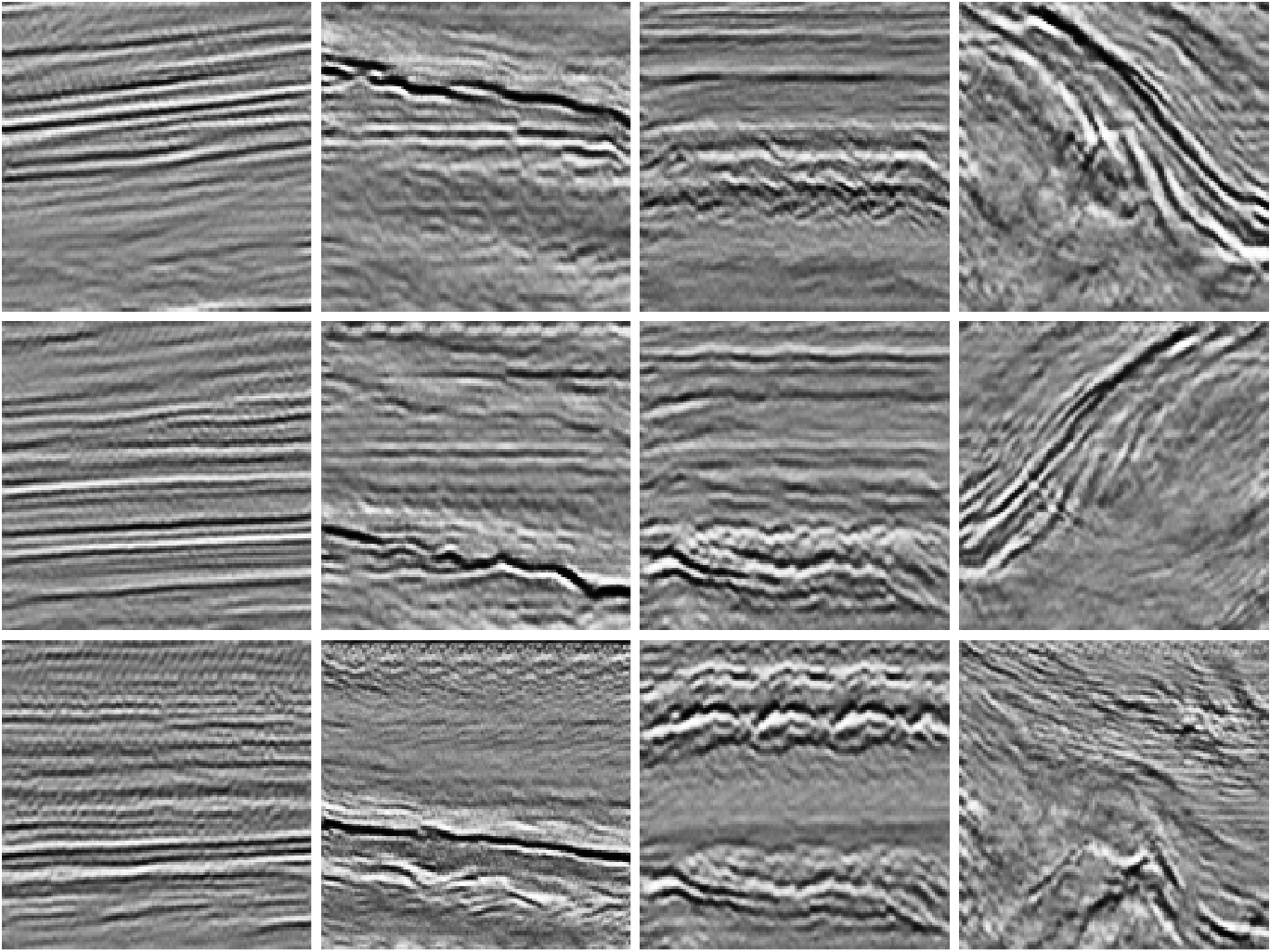}
	\caption{Sample seismic images generated using DCGAN. Each column in the figure contains generated images that represent one of the four classes of \texttt{LANDMASS} datasets, i.e. \texttt{Horizons}, \texttt{Faults}, \texttt{Chaotic horizons}, and \texttt{Salt domes}, respectively.}
	\label{fig:GAN}
\end{figure}

\textit{Generative Adversarial Networks}: Among various deep learning techniques, generative models are a set of models that have been quite popular recently. Generative models are systems that can learn and generalize the probability distribution of data from training samples. The outcome of such systems can be either the explicit probability distribution or samples from the learned models. Some models can be designed to do both tasks. One specific type of generative models that have gained popularity recently is the generative adversarial networks (GAN). GANs are used to generate samples that come from the same distribution as some given training samples. The task is achieved by designing two separate systems that compete against each other. More specifically, GANs consist of: 1) a \emph{generator} that can transform random noise into an image that looks like the training samples, and 2) a \emph{discriminator} that takes an image as input and estimates the probability that this image is from the same distribution as that of the training samples. After proper training, the generator will be able to produce images that resemble images from the training samples. GANs have been employed in various applications such as image synthesis, image interpolation and inpainting, style transfer, and next-frame prediction \cite{zhu2017CycleGAN, ledig2016SRGAN, radford2015DCGAN, lotter2015GAN}. More details about GANs and their useful applications can be found in \cite{goodfellow2016nips}. However, in the context of seismic interpretation, we highlight three applications of GANs that are particularly useful. The first application is to use GANs to train a model with unlabeled or partially labeled data (e.g. \cite{GAN2015unsupervised}). A second application is model-based seismic interpolation using GANs to interpolate seismic traces or generate new seismic sections at super-resolution. A third application is to utilize GANs for style transfer of seismic datasets, which are very diverse in terms of acquisition resolution, reflector strength, and size of structures. Due to the diversity of such datasets, an algorithm trained on one seismic dataset will not necessarily work on other datasets. Using GANs, it is possible to transfer the style of one dataset along with its labels to another dataset that can be used to improve the robustness of the algorithm on a different dataset. Fig.~\ref{fig:GAN} shows sample seismic images generated by a deep convolutional generative adversarial network (DCGAN) \cite{radford2015DCGAN}. The samples are arranged such that each column contains images of the same class. The network was trained on the entire LANDMASS dataset \cite{LANDMASS}. 
\\

\textit{Active Learning}: In a typical interpretation process, an interpreter works in cycles of interpretation-examination to improve the results gradually. It is desirable if computational approaches can mimic this interactive and iterative process. Among all varieties of machine learning techniques, active learning emerges as a perfect fit to fulfill this purpose. The inherent interactivity will involve a user to assist in making intermediate decisions, guiding the learning process to follow the correct path. More importantly, active learning will also help alleviate the problem of insufficient manually labeled data, as it adds samples to the set of labeled data during each iteration. A typical active learning process starts with a limited set of manually labeled data samples. Initial models are trained using the limited labeled data and then used to select a subset from a much larger set of unlabeled samples. The chosen samples are provided as queries to an interactive annotation procedure, where they are labeled by a human annotator (or a group of annotators to reduce the chance of a biased decision). Carefully designed computer algorithms may also be employed to perform the labeling in place of human annotators. Finally, the labeled queries are added to the existing labeled data set, with which a new iteration of learning can be performed. For active learning, the query selection is the key to the whole process and the most active area of current research.\\

\textit{Attention Models Inspired by the Human Visual System}: In addition to machine learning, we believe the other important factor that will impact seismic interpretation research is the HVS modeling. For visual analysis, HVS reduces the amount of the sensory data information, also known as the task-free visual search, by focusing on the \emph{perceptually} salient segments of visual data that conveys the most useful information about the scene. For example, as an interpreter traverses through a seismic volume or a video, the HVS that is sensitive to structures and variations in amplitude and surrounding environment registers the relative motion of seismic structures across all three dimensions, not all of which are characterized by strong seismic reflections or texture contrast. Therefore, computational algorithms inspired by HVS can be used to develop attention models that can effectively automate the structural interpretation of seismic volumes. Among many, one of such attention models is defined as a saliency model, which attempts to predict the interesting areas in images and videos, typically called salient regions, by relying on low-level visual cues. In seismic interpretation, visual saliency is important to predict the attention of human interpreters and highlight areas of interest in seismic sections, which can not only help in the automation of the interpretation process but also develop attention models that can be applied to new datasets. Recently, a seismic attribute called SalSi is proposed in~\cite{Shafiq_ICASSP2016, Shafiq2017_Sal_GP}, which highlights salient areas within seismic volumes to assist interpreters in structural interpretation of salt domes and faults. Furthermore, by incorporating a priori information into saliency detection, attention models can be used to mimic the behavior of interpreters looking at seismic sections. Therefore, saliency detection is one of the promising directions that can not only assist interpreters in the interpretation process but also help develop better attention models for seismic interpretation. \\

It is a challenging task to train a machine to understand and reveal various structures of Earth's subsurface. Yet, once achieved, it will be a truly rewarding accomplishment that has a large impact on the economy and society. Through its fast advancement in the past decade, seismic interpretation has already demonstrated its high potential in fulfilling this mission. With a broader and deeper collaboration between the geophysical and the signal processing communities, we expect to witness more breakthroughs in the near future in seismic-interpretation-enabled subsurface understanding.

\section*{Acknowledgments}

The authors would like to acknowledge the support of the Center for Energy and Geo Processing (CeGP) at the Georgia Institute of Technology and King Fahd University of Petroleum and Minerals (KFUPM, Project GTEC 1401-1402).

\ifCLASSOPTIONcaptionsoff
  \newpage
\fi



%

\bibliographystyle{IEEEtran}
\bibliography{main.bib}

%





\end{document}